\documentclass[journal]{IEEEtran}

\usepackage{caption}
\usepackage{subcaption}
\ifCLASSINFOpdf
  \usepackage[pdftex]{graphicx}
  \graphicspath{{./graph/}}
  \DeclareGraphicsExtensions{.pdf,.jpeg,.png}
\else
  \usepackage[dvips]{graphicx}
  \graphicspath{{./graph/}}
  \DeclareGraphicsExtensions{.eps}
\fi

\ifCLASSINFOpdf
\else
\fi

\usepackage[cmex10]{amsmath,mathtools}
\usepackage{mdwmath}
\usepackage{mdwtab}
\usepackage{amssymb}
\usepackage{tabularx}
\usepackage{float}
\usepackage{multirow}
\usepackage{color}
\usepackage{multicol}

\usepackage{algorithmic}
\usepackage{algorithm}
\usepackage{float}

\newtheorem{assumption}{Assumption}
\newtheorem{theorem}{Theorem}

\allowdisplaybreaks[4]
\makeatother
\usepackage{epstopdf}
\usepackage{color}

\usepackage[utf8]{inputenc}

\usepackage{cite}

\begin{document}

\title{Data-Agnostic Model Poisoning against Federated Learning: A Graph Autoencoder Approach}

\author{Kai~Li,~\IEEEmembership{Senior Member,~IEEE,}
        Jingjing~Zheng,~\IEEEmembership{Student Member,~IEEE,}
        Xin~Yuan,~\IEEEmembership{Member,~IEEE,}
        Wei~Ni,~\IEEEmembership{Senior Member,~IEEE,}
        Ozgur~B.~Akan,~\IEEEmembership{Fellow,~IEEE,}
        and~H. Vincent Poor,~\IEEEmembership{Life Fellow,~IEEE}
\thanks{K.~Li is with the Division of Electrical Engineering, Department of Engineering, University of Cambridge, CB3 0FA Cambridge, U.K., and also with Real-Time and Embedded Computing Systems Research Centre (CISTER), Porto 4249--015, Portugal (E-mail: kaili@ieee.org).}
\thanks{J.~Zheng is with CyLab Security and Privacy Institute, Carnegie Mellon University, Pittsburgh, PA 15213, USA, and also with Real-Time and Embedded Computing Systems Research Centre (CISTER), Porto 4249--015, Portugal (E-mail: zheng@isep.ipp.pt).}
\thanks{X.~Yuan and W.~Ni are with the Digital Productivity and Services Flagship, Commonwealth Scientific and Industrial Research Organization (CSIRO), Sydney, NSW 2122, Australia (E-mail: \{xin.yuan,wei.ni\}@data61.csiro.au).}
\thanks{O. B. Akan is with the Division of Electrical Engineering, Department of Engineering, University of Cambridge, CB3 0FA Cambridge, U.K., and also with the Center for NeXt-Generation Communications (CXC), Ko\c c University, 34450 Istanbul, Turkey (E-mail: oba21@cam.ac.uk).}
\thanks{H. V. Poor is with the Department of Electrical and Computer Engineering, Princeton University, Princeton, NJ 08544, USA (E-mail: poor@princeton.edu).}
}


\IEEEcompsoctitleabstractindextext{%
\begin{abstract}
\boldmath
This paper proposes a novel, data-agnostic, model poisoning attack on Federated Learning (FL), by designing a new adversarial graph autoencoder (GAE)-based framework. The attack requires no knowledge of FL training data and achieves both effectiveness and undetectability. By listening to the benign local models and the global model, the attacker extracts the graph structural correlations among the benign local models and the training data features substantiating the models. The attacker then adversarially regenerates the graph structural correlations while maximizing the FL training loss, and subsequently generates malicious local models using the adversarial graph structure and the training data features of the benign ones. A new algorithm is designed to iteratively train the malicious local models using GAE and sub-gradient descent. The convergence of FL under attack is rigorously proved, with a considerably large optimality gap. Experiments show that the FL accuracy drops gradually under the proposed attack and existing defense mechanisms fail to detect it. The attack can give rise to an infection across all benign devices, making it a serious threat to FL.
\end{abstract}

}

\maketitle

\IEEEdisplaynotcompsoctitleabstractindextext
\IEEEpeerreviewmaketitle

\section{Introduction}
\label{sec_intro}
The use of mobile edge computing is increasingly prevalent, especially in catering to user devices that come with a multitude of sensors. These sensors produce vast amounts of data, like images recording human activities or the real-time locations of vehicles, as seen in smart city scenarios~\cite{zhang2022lsfl}. However, transferring this training data from the user's device to a server can pose a threat to data privacy leakage. Federated Learning (FL) is an emerging distributed machine learning approach that gains traction as a solution to mitigate data privacy concerns~\cite{khan2021federated}. With FL, user devices can jointly train a machine learning model without having to disclose their private data to a server. 
The user devices, acting as clients, iteratively train their local models on their private data and send the local model updates to a server. 
At the server, a global model is updated without collecting private data from the user devices. The global model is then sent back to the user devices, allowing them to continue training their local models based on the global model and their local data~\cite{xu2022laf}. This process helps to support data privacy and allows for real-time processing capabilities at the edge of networks, making FL a significant aspect of mobile edge computing~\cite{zhou2022pflf,zhou2022privacy}. 

Despite the fact that FL can help prevent attackers from accessing the private data of user devices, an attacker (in most cases, a malicious user device) can potentially launch model poisoning or data poisoning attacks to manipulate FL and propagate the attacks into benign user devices~\cite{jere2020taxonomy,rahman2020internet}, resulting in a failure of FL training. Specifically, model poisoning aims to send malicious local model updates to the server during an aggregation process. The malicious update can introduce specific vulnerabilities in the global model or simply degrade FL performance. By contrast, data poisoning attempts to inject malicious data or modify existing data on user devices to misguide local model training, thus compromising local model updates. Existing data poisoning attacks generally require an attacker to have some knowledge of the datasets used for FL training~\cite{tian2022comprehensive}, so that it can extract and manipulate the features of the datasets for effective attacks~\cite{zhang2022fldetector}. By launching model poisoning attacks~\cite{fang2020local} or data poisoning attacks~\cite{lyu2020threats}, an attacker could manipulate either the hyperparameters of the local models or the training datasets of benign users to compromise learning accuracy.

Much less constrained and potentially more threatening model poisoning attacks on FL would result if they could be based solely on the benign local models overheard by an attacker and the global models broadcast by the aggregator; i.e., when the attacker has no access to the training data. However, without training data, it is challenging for the malicious local models to strike a balance between effectiveness and undetectability~\cite{shejwalkar2022back}. To the best of our knowledge, such attacks are new and have not been reported in the literature. 

In this paper, we propose a new, data-agnostic, model poisoning attack on FL systems, where an adversarial graph autoencoder (GAE)~\cite{cai2018comprehensive} is designed to generate malicious local models solely based on the benign local models overheard and capturing the correlation features of the benign local and global models. Specifically, an attacker overhears the benign local models uploaded by the user devices, and the global model broadcast by the server. It extracts the graph structure capturing the correlations between the benign local models, and decouples the graph structure from underlying data features substantiating the local models.
The attacker first regenerates manipulatively the graph structure to retain the structural features of the local models and maximize the FL training loss by using the GAE, and then generates malicious local models by applying the regenerated graph structure to the data features of the benign local models. As a result, the malicious local models can effectively compromise the global model, while remaining compatible with the benign models and hence reasonably undetectable. 

The contributions of the paper are summarized below.
\begin{itemize}
    \item 
    A new design of data-agnostic, malicious local models, which manipulates the correlations of benign local models and retains the genuine data features substantiating the benign local models; 
    \item
    A new GAE framework, which is trained together with sub-gradient descent to regenerate manipulatively the correlations of the local models while keeping the malicious local models undetectable; and 
    \item
    A rigorous analysis, which proves the convergence of the global model under attack, but to an inferior optimality gap.
\end{itemize}
Extensive experiments indicate that the FL accuracy drops gradually under the proposed attack, and the existing poisoning defense mechanisms can hardly detect the attack. Since the malicious local models are uploaded to the server for global model aggregation, the proposed attack gives rise to an epidemic infection across all benign devices.

The proposed GAE-based attack on FL involves attackers intentionally poisoning malicious local models, aiming to degrade or manipulate the performance of the global model. The attack challenges the security, privacy, and robustness of FL. While security is threatened by unauthorized access or malicious insiders tampering with local models, privacy concerns arise when the attackers try to reverse-engineer or glean information about the benign devices' data. Moreover, robustness, which is the ability of FL to consistently produce reliable and accurate results, can be directly undermined, as poisoned local models compromise the integrity and efficacy of FL. To this end, the proposed GAE-based attack poses a comprehensive threat to the security, privacy, and robustness of FL.

The rest of this paper is organized as follows. Section~\ref{sec_relatedwork} introduces the background of adversarial attacks against wireless systems and FL. Section~\ref{sec_systems} discusses FL with benign user devices and server, as well as the eavesdropping model. The proposed GAE-based epidemic attack is delineated in Section~\ref{sec_GAE}. Performance analysis is conducted in Section~\ref{sec_evaluation}. Section~\ref{sec_cond} concludes the paper. 

\section{Related Work}
\label{sec_relatedwork}
This section reviews the literature on adversarial attacks against wireless systems as well as FL, including model and data poisoning attacks. 
On the one hand, because of their broadcast nature, wireless channels are particularly vulnerable to eavesdropping attacks. An attacker is likely to overhear the local model updates transmitted by the other benign users in wireless FL. On the other hand, the model poisoning attack considered in this paper has not been studied in the literature. Instead, existing attacks on wireless FL have focused primarily on building an adversarial data classification/label model for attackers, according to the data packets and features overheard, e.g.,~\cite{zhao2022pvd} and~\cite{zhou2021hierarchical}. There is clearly an opportunity for the new attack to strike.

\subsection{Adversarial Attacks on Wireless Systems}
In~\cite{singh2021adversarial}, an adversarial attack was studied to manipulate the measurement of smart meters in residential homes. Smart meter data could inform residents of which appliances consumed the most electricity and adjust energy production. The attacker employed deep learning to train a power usage pattern classification model and generated malicious data that was indistinguishable from the true data.
In~\cite{zhao2022pvd}, machine learning was used to generate an adversarial attack for targeting data fusion or aggregation. The attacker infiltrated some devices and learned the decision process and data fusion settings by observing data exchanges between the devices and the data center.

In~\cite{bao2021threat}, the authors analyzed targeted adversarial attacks that aimed to manipulate the output of a convolutional neural network (NN)-based classifier. They also evaluated non-targeted adversarial attacks against convolutional NN-based device identification. To evaluate these attacks, the authors used combined indicators of logits to increase the perturbation levels and iterative steps, resulting in a high success rate of adversarial attacks.
In~\cite{rahman2020adversarial}, researchers used deep learning to recognize COVID-19 symptoms by training on medical data from user devices. They evaluated several adversarial attacks that aimed to falsify the data and symptom recognition. The study found that existing deep learning algorithms were vulnerable to these attacks, highlighting the need for advanced security measures.

In~\cite{zhou2021hierarchical}, an adversarial attack was developed to deactivate graph-based intrusion detection in a targeted wireless system. The attack began by building a shadow graph based on overheard data packets and features. A random walk algorithm was then used to evaluate each node in the attacker's graph, selecting the node with the largest weight to attack. The attack would perturb data features and alter classification labels. In~\cite{abusnaina2019adversarial}, an adversarial attack was developed to utilize graph embedding and augmentation to misclassify system malware samples as benign. The graph-based attack aimed to embed a target malware sample into benign software. By combining the benign code sample and the target malware sample in the graph, the adversarial attack could learn complex features, resulting in a high misclassification rate at the user device.

In~\cite{talpur2021adversarial}, a study was conducted on a Sybil-based data poisoning attack against deep reinforcement learning-based service placement in the Internet of Vehicles (IoV). The attack targeted the agent that is responsible for learning the service quality and deciding on service placement based on delay. A Sybil attacker, which is a malicious vehicle, used data poisoning techniques to masquerade as a legitimate vehicle by stealing or borrowing its identity. The attacker then maliciously sent false data to other vehicles.

Unfortunately, it is difficult for the attacker to formulate the adversarial data classification/label model in FL systems since the benign user devices can collaboratively conduct model training without sharing their private data. 

\subsection{Poisoning Attacks on FL}
In order to corrupt the FL, the attacker can launch either a data poisoning or a model poisoning attack. In the data poisoning attack, the attacker injects fake data with manipulated features and flips labels into the benign user devices. In the model poisoning attack, the attacker submits malicious local models to the server. Both attacks aim to corrupt the FL by introducing false information.

In~\cite{shejwalkar2022back}, the authors systematically categorized the existing threat models associated with poisoning attacks on FL, where practical boundaries of numerous parameters pertinent to FL robustness were delineated. An array of untargeted model and data poisoning attacks on FL was analyzed to encompass the existing attack strategies. A model poisoning attack was developed using gradient ascent to fine-tune the global model and increase its loss on benign data. The model poisoning attack adjusts the $L_2$-norm of the poisoned model update to circumvent the robustness criterion of the model aggregation.

In~\cite{chen2020zero}, an adversarial attack mitigation scheme based on clustering was studied. The scheme aimed to protect FL by using unsupervised weight training to split and merge weight clusters at the server to filter out malicious local models that were uploaded by the user devices without identity verification. In~\cite{tolpegin2020data}, malicious local models were derived from mislabeled data to manipulate the global model. The study found that this attack could result in a significant drop in classification accuracy, and that it was difficult to detect due to its negative impact on the target device and minimal impact on other benign devices.

In~\cite{gao2021secure}, an inference model was formulated to take local models as input and output the categories of data. A malicious local model based on a differential selection strategy was used to select two adjacent categories. To approximate the benign local model, a category inference attack was studied, in which the attacker learns the data features underlying benign local models. 

The authors of~\cite{wang2020attack} presented a backdoor attack against FL in mobile edge computing (MEC), which targeted the tail of the input data distribution at the local devices. The attack used projected gradient descent to maintain the distance between the malicious local model and the global model, to misclassify the targeted samples and bypass defense mechanisms. 

In~\cite{zhang2020poisongan}, generative adversarial networks (GANs) were utilized to construct data poisoning attacks against FL. The attacker trained the GAN to replicate the local data of the benign devices. Since the attacker had no information about the local data, the GAN-based data poisoning updated the global model to re-select the potential targeted devices. In~\cite{zhang2019poisoning}, a GAN-based FL poisoning attack was studied, where the attacker posed as one of the benign devices and trained the GAN to mimic the dataset of the benign devices. The malicious data generated by the attacker were trained to compromise the global model. In~\cite{wang2019beyond}, a malicious server deployed a GAN-based reconstruction attack against FL to tamper with the private data of the user devices. The malicious server discriminated the devices' identities and data representatives to supervise the training of GANs and generate malicious data for each specific device. In~\cite{song2020analyzing}, the authors focused on a device-level privacy leakage attack launched by a malicious server. A GAN-based framework was presented to discriminate the data category and device's identity and recover the private data of the device. 
The attack could associate the data features from different devices to re-identify the local models.

Unfortunately, the existing data poisoning or model poisoning attacks have not exploited the implicit relationship between local models~\cite{caldarola2021cluster,mei2019sgnn}. Moreover, the existing poisoning attacks generally require the attacker to have the knowledge of (part of) the datasets used for FL training.

\section{System Model}
\label{sec_systems}
In this section, we first describe an FL training process, e.g., for image classification. Next, we present the threat model, where malevolent devices can act as attackers. An attacker creates and uploads malicious local model updates to progressively contaminate the global model of the FL. At last, we describe an attacker detection model that the server can adopt to discern malicious local models by measuring the Euclidean distances between the models.

\subsection{Federated Learning}
We assume there are $J$ benign user devices and an authorized (legitimate) but malicious user device (or an attacker) in the FL training process. 
A benign user device $j \in [1, J]$ has $D_j(\tau)$ amount of data at the $\tau$-th iteration. Let $x_j^i$ and $y_j^i$ denote the input of the captured images and the output of the FL model at device $j$, respectively. $i \in [1, D_j(\tau)]$. A training loss function of device $j$, denoted by $f_i(\pmb{\omega}_j(\tau); x_j^i, y_j^i)$, captures approximation errors over the input $x_j^i$ and the output $y_j^i$. Here, $\pmb{\omega}_j(\tau)$ is the weight parameter of the loss function in the model being trained by the FL. For instance, $f_i(\pmb{\omega}_j(\tau); x_j^i, y_j^i)$ can be modeled by linear regression, i.e., $f_i(\pmb{\omega}_j(\tau); x_j^i, y_j^i) = \frac{1}{2} (\pmb{\omega}_j(\tau)^T {x_j^i} - y_j^i)^2$; or logistic regression, i.e., $f_i(\pmb{\omega}_j(\tau); x_j^i, y_j^i) =  y_j^i \log \Big(1 + \exp \big(-\pmb{\omega}_j(\tau)^T {x_j^i}\big)\Big) - (1- y_j^i) \log \Big(1-\frac{1}{1 + \exp \big(-\pmb{\omega}_j(t)^T {x_j^i} \big)}\Big) $. Here, $(\cdot)^T$ denotes transpose.
Given $D_j(\tau)$, the local loss function of the FL at device $j$ for the $\tau$-th iteration is  
\begin{align}
F_j\!(\pmb{\omega}_j(\tau))\!=\! \frac{1}{D_j(\tau)} \sum_{i=1}^{D_j\!(\tau)}\! f_i(\pmb{\omega}_j(\!\tau\!);\! x_j^i, \!y_j^i\!) \!+\! \mu g(\pmb{\omega}_j(\!\tau\!)\!), 
\label{eq_lossFunc}
\end{align}
where $g(\cdot)$ is a regularizer function that represents the effect of the local training noise, and $\mu \in [0,1]$ is a coefficient~\cite{guo2020v}.

The local model of user device $j$ is updated by
\begin{equation}
\pmb{\omega}_j(\tau +1) = \pmb{\omega}_j(\tau) - \eta \nabla F_j (\pmb{\omega}_j(\tau)),
\label{eq_local_SGD}
\end{equation} 
where $\eta$ is the learning rate. 

After every $T_L$ local updates (or iterations), there is a communication round where the benign user devices upload their local models to a server. The server aggregates the local models to update the global model and broadcasts the global model to all user devices. 

\subsection{Threat Model}
\begin{figure}
\centering
\includegraphics[width=3.3in]{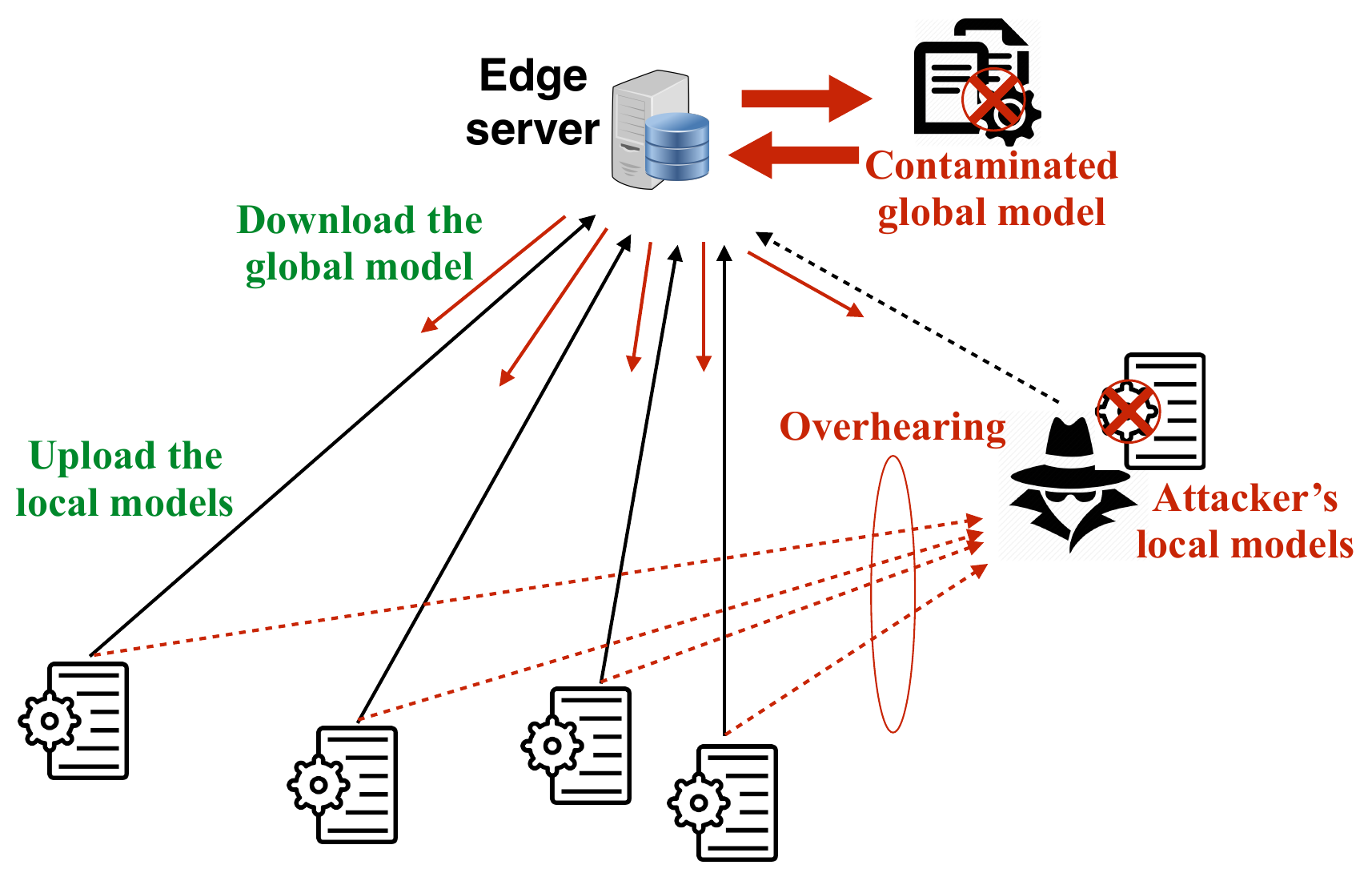}
\caption{The proposed data-agnostic, model poisoning attack, where the attacker overhears the global model and the local models uploaded by the benign user devices. Next, the attacker generates a malicious local model to contaminate the global model and the benign local models.} 
\label{fig_attackingFL}
\end{figure}

We consider a new data-agnostic model poisoning attack, where malicious local models are generated solely based on the benign local models overheard and the correlation features of the benign local and global models. This attack could be particularly severe in FL systems under wireless settings, due to the broadcast nature of radio. As shown in Fig.~\ref{fig_attackingFL}, an attacker within the vicinity of benign user devices and equipped with radio transceivers can passively eavesdrop on the transmitted local models of some (if not all) of the benign user devices, extracting their features and generating its own malicious local model. Although cryptography can prevent eavesdropping attacks to some extent, existing techniques, such as those developed in~\cite{aviram2016drown,hebrok2023we,diaz2019tls}, have demonstrated the possibility of deciphering encrypted information with limited initial data.

The attacker creates and uploads a malicious local model, denoted by $\pmb{\omega}^a(t)$, to contaminate the global model $\pmb{\omega}_g(t)$, and subsequently the local models of the benign users, i.e., $\pmb{\omega}_j(t)$, $\forall j \in [1, J]$, where $t$ indicates the $t$-th communication round. $\pmb{\omega}^a(t)$ is adversarially created based on the benign local model parameters overheard by the attacker in the $t$-th communication round.


Unaware of the ill-intentioned attacker, the server aggregates the local models of all user devices, including both the benign and malicious local models, and unintentionally creates a contaminated global model, denoted by $\pmb{\omega}_g^a(t)$, at the $t$-th communication round. The total size of the local training data reported to the server is $D(t) = \sum_{j=1}^{J} D_j(t)+D_a(t)$, where $D_a(t)$ is the claimed data size of the attacker at the $t$-th communication round. Then, the contaminated global model is given by 
\begin{align}
\pmb{\omega}_g^a(t) =  \sum_{j=1}^J  \frac{D_j(t)}{D(t)}\,\pmb{\omega}_j(t) + \frac{D_a(t)}{D(t)}\pmb{\omega}^a(t),
\label{eq_glbAttacks}
\end{align}
The server broadcasts $\pmb{\omega}_g(t)$ to all user devices.

To this end, the FL training process in essence trains the global model based on the local datasets of all user devices, including the nonexistent dataset claimed by the attacker, by minimizing the following global loss function:
\begin{equation}\label{eq_minGlobalLoss}
\underset{\pmb \omega^a_g  (t)}{\min} F({\pmb \omega^a_g} (t))\!=\!  \sum_{j=1}^J \frac{D_j(t)}{D(t)} F_j(\pmb{\omega}^a_g(t))\!+ \!\frac{D_a(t)}{D(t)}F_a(\pmb{\omega}^a_g(t)),
\end{equation}
where $F_a(\cdot)$ is the claimed local loss function of the attacker, which is claimed to conform to~\eqref{eq_lossFunc}.

To attack the FL training process, the attacker aims to maximize $F({\pmb \omega^a_g} (t))$, while keeping $\pmb{\omega}^a(t)$ undetectable by the server that typically constantly assesses the similarities among all local models and rules out those substantially different from the rest, e.g., Krum or multi-Krum~\cite{blanchard2017machine}. As a result, the attacked global model diverges in a direction opposite to the one intended in the absence of the attack. 

\begin{figure*}[htb]
\centering
\includegraphics[width=6.7in]{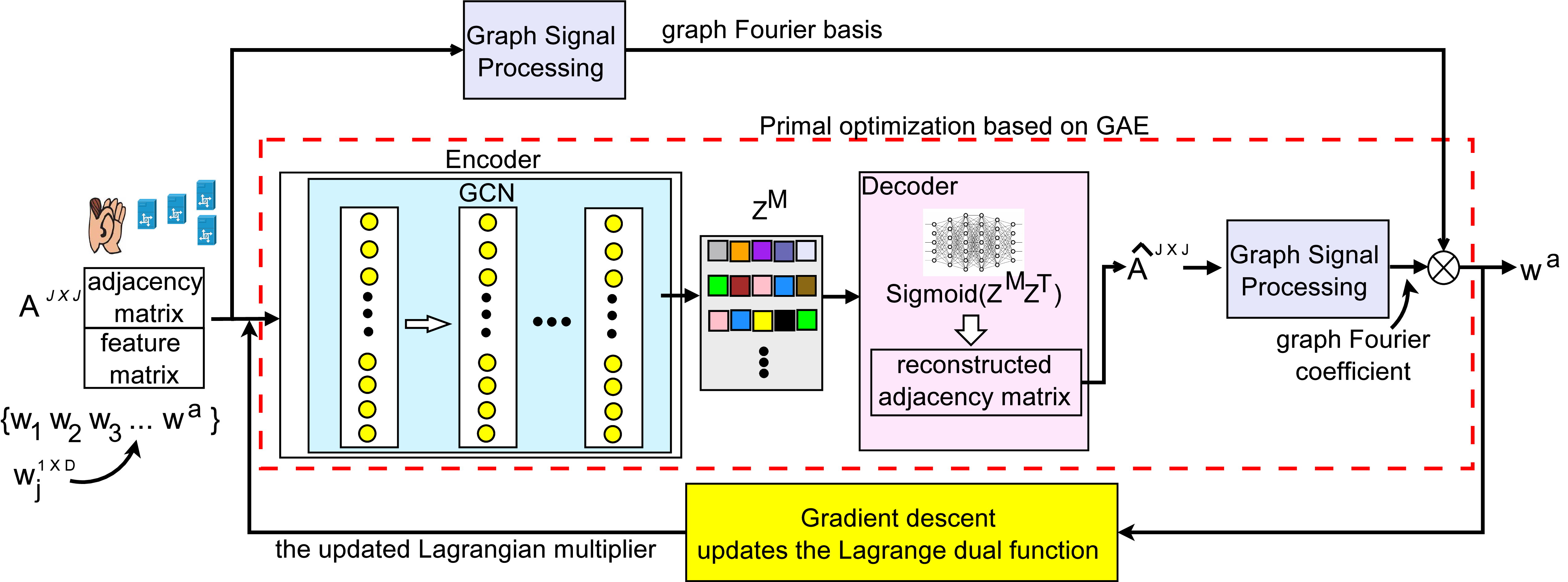}
\caption{The proposed GAE model for generating { data-agnostic, malicious local models,} where the attacker overhears $\pmb{\omega}_g(t)$ and $\pmb{\omega}_j(t),\,\forall j$ and applies the GCN-based encoder to create $\mathcal{Z}^M$. The output of the encoder, i.e., the feature representations, is input to the decoder for feature reconstruction. }
\label{fig_graph}
\end{figure*}

At the $t$-th communication round, the attacker formulates a data-agnostic, model poisoning attack problem:
\begin{subequations}\label{eq_opt}
\begin{align}
~~\max_{\pmb{\omega}^a(t)} & ~~F(\pmb{\omega}_g^a(t))\label{eq_opt_obj}\\
{\rm s.t.} &~~d(\pmb{\omega}^a(t), \pmb{\omega}_g^a(t)) \leq d_T, \label{eq_const_dist}
\end{align}
\end{subequations}
where $d(\pmb{\omega}^a(t), \pmb{\omega}_g^a(t))$ evaluates the Euclidean distance between $\pmb{\omega}^a(t)$ and $\pmb{\omega}_g^a(t)$, and $d_T$ is a pre-specified threshold that ensures the generated malicious local model is close to the global model in the Euclidean space to escape the scrutiny of the server.

\subsection{Defense Model for Attacker Detection}
In response to the prevalent threat of model poisoning in FL, an attacker detection model residing on the server can be applied, which leverages the Euclidean distance metric to discern malicious local models, for instance,~\cite{zhang2022fldetector} and~\cite{li2021lomar}. By measuring the straight-line distance between each incoming local model and the aggregated global model, this model aims to identify anomalous deviations indicative of malicious intent. The underlying rationale is that genuine local models from benign devices are expected to cluster within a certain proximity in the model space, while malicious local models, designed to sabotage the global model's integrity, would exhibit more pronounced deviations. By setting a distance threshold, local models that exceed this threshold can be flagged or discarded, effectively isolating and mitigating the impact of malicious local models on the global model's integrity. This server-side defense mechanism underscores the potential of geometric measures, like Euclidean distance, as powerful tools in safeguarding FL systems from adversarial attacks.

\section{Proposed Data-Agnostic Model Poisoning Attack on FL}
\label{sec_GAE}
In this section, we elaborate on the proposed data-agnostic model poisoning attack, where adversarial GAE is designed to extract the feature correlation among the local models of the benign user devices and reconstruct an adversarial adjacency matrix. With the adjacency matrix, the attacker trains the GAE to generate malicious local models without being detected by the server.

\subsection{GAE Model for Data-Agnostic Model Poisoning}
The arbitrary features of $\pmb{\omega}^a(t)$ and those of the benign local models may have a low feature correlation, which can be potentially detected by the server. To address this, we develop a new GAE model for the novel, data-agnostic, model poisoning attack.

The optimization problem in~\eqref{eq_opt} can be transformed using the Lagrangian method~\cite{tran2021differentially}. Let $\lambda$ denote the dual variable. The Lagrange function is given by 
\begin{align}
L(\pmb{\omega}^a(t), \lambda) = &F(\pmb{\omega}_g^a(t)) + \lambda (d_T - d(\pmb{\omega}^a(t), \pmb{\omega}_g^a(t))).
\label{eq_lag}
\end{align}
The Lagrange dual function is
\begin{equation}
{\cal D}(\lambda) = \max_{\pmb{\omega}^a(t)} L(\pmb{\omega}^a(t), \lambda).
\end{equation}
The dual problem of the problem in~\eqref{eq_opt} is given by 
\begin{equation}\label{eq-dual}
\min_{\lambda(t)} {\cal D}(\lambda).
\end{equation}
At the $t$-th communication round, given $\lambda=\lambda(t)$, the primary variable $\pmb{\omega}^a(t)$ of the data-agnostic model poisoning attack can be optimized by solving
\begin{equation}
{\pmb{\omega}^a(t)}^* = \arg \max_{\pmb{\omega}^a(t)} \{F(\pmb{\omega}_g^a(t)) -
\lambda(t) d(\pmb{\omega}^a(t), \pmb{\omega}_g^a(t))\}.
\label{eq_dual_equation}
\end{equation}
With obtained ${\pmb{\omega}^a(t)}^*$,  the sub-gradient descent method can be taken to update $\lambda(t)$ by solving the dual problem~\eqref{eq-dual}. Specifically, $\lambda(t)$ is updated by~\cite{boyd2004convex}
\begin{equation}\label{eq-sub-gradient}
\lambda\left(t +1 \right)  = \left[ \lambda(t ) - \varepsilon\left( d(\pmb{\omega}^a(t)^*, \pmb{\omega}_g^a(t)) - d_T \right) \right]^+,
\end{equation}
where $\varepsilon$ is the step size, $\tau$ is the index to the iterations, and $\left[x \right]^+ = \max\left(0,x \right)$. At initialization, $\lambda(t)$ is non-negative, i.e., $\lambda(1) \geq 0$, to ensure \eqref{eq-sub-gradient} converges. 

We propose to solve~\eqref{eq_dual_equation} by developing a new GAE model, followed by the sub-gradient descent to update \eqref{eq-sub-gradient}. These two steps are performed in an alternating manner, as illustrated in Fig.~\ref{fig_graph}. 
Specifically, we propose to decompose the local model parameters of the benign devices into a graph capturing the correlations (or similarity) between the benign local models, and the underlying spectral-domain data features that the local models capture. Then, we regenerate the graph with the GAE in a manipulative manner and subsequently compose malicious local models with the regenerated graph and the original, genuine data features. The rationale of this design is provided as follows. 
\begin{itemize}
\item By regenerating the graph with the GAE, we retain and manipulate the correlations between the local models, and also deter the convergence of the global model, i.e., by maximizing \eqref{eq_dual_equation}. The decoder of the GAE reproduces the correlations while satisfying constraint~\eqref{eq_const_dist}. This suppresses structural dissimilarity between the malicious and benign local models. 
\item By using the genuine underlying spectral-domain data features, the malicious local models are substantiated by the genuine data features. Hence, they are less likely to be detected by the server. 
\end{itemize}

\subsubsection{GAE for Malicious Model Generation}
The attacker aims to construct $\pmb{\omega}^a(t)$ without knowing any data of the benign devices. As illustrated in Fig.~\ref{fig_graph}, a graph, denoted by $G({{\cal V}}, E, \mathcal{F})$, is used to formulate the benign local models in FL, where {${\cal V}$}, $E$, and $\mathcal{F}$ represent vertexes, edges, and the feature matrix of the graph, respectively. 

Let $\mathcal{F} = [\pmb{\omega}_1(t), \cdots, \pmb{\omega}_j(t), \pmb{\omega}^a(t)]$ collect all local models of both benign and malicious devices. $\pmb{\omega}_j(t),\pmb{\omega}^a(t) \in \mathbb{R}^{1 \times D},\,\forall j$. Also, let $\mathcal{A} \in \mathbb{R}^{J \times J}$ denote the adjacency matrix that describes the correlation among the local models of the user devices. At the $t$-th communication round of the FL, the $(j, j^\prime)$-th element of $\mathcal{A}$, denoted by $\overline{\omega}_{j,j^\prime}$ ($j, j^\prime \in [1, J]$), measures the inner product between $\pmb{\omega}_j(t)$ and $\pmb{\omega}_{j^\prime}(t)$~\cite{zhu2020anomaly}, as given by
\begin{equation}
\overline{\omega}_{j,j^\prime} = \frac{\pmb{\omega}_j(t) \cdot \pmb{\omega}_{j^\prime}(t)}{\|\pmb{\omega}_j(t)\|\cdot \|\pmb{\omega}_{j^\prime}(t)\|}. 
\label{eq_innerProduct}
\end{equation}
According to $\mathcal{A}$, the topological structure of the graph $\mathcal{G}$ can be constructed.

The GAE consists of an encoder and a decoder, where the encoder encodes the graph data with the features and the decoder takes the encoder's output as the input to reconstruct $G({{\cal V}}, E, \mathcal{F})$~\cite{wang2020simple}. 

$\bullet$ \textbf{Encoder: }
The encoder in the proposed GAE is responsible for mapping $G({{\cal V}}, E, \mathcal{F})$ to a lower-dimensional representation. We build the encoder based on an $M$-layer graph convolutional network (GCN) architecture, which learns a representation that captures the underlying features of $G({{\cal V}}, E, \mathcal{F})$. The encoded representation is then used as input to the decoder, to reconstruct the original graph from the lower-dimensional representation to obtain the malicious local model ${\pmb{\omega}^a(t}^*)$ in~\eqref{eq_dual_equation}.

The encoder takes $\mathcal{A}$ as its input to its $M$-layer GCN. The output at the $M$-th layer is 
\begin{align}
\mathcal{Z}^M = f_G(\mathcal{Z}^{M-1}, \mathcal{A} | \textbf{w}^M), 
\label{eq_Z}
\end{align}
where $f_G(\cdot,\cdot|\cdot)$ is a spectral convolution function and $\textbf{w}^M$ defines the weight matrix at the $M$-th layer of the GCN.

With the identity matrix $I \in \mathbb{R}^{J\times J}$, we define $\widetilde{\mathcal{A}} = \mathcal{A} + I$ and $\overline{\mathcal{A}}_{jj} = \sum_{j^\prime} \widetilde{\mathcal{A}}_{jj^\prime}$. To generate a feature representation of the graph, the encoder can be written as 
\begin{align}
f_G(\mathcal{Z}^{M-1}, \mathcal{A} | \textbf{w}^M) = \Phi^M (\overline{\mathcal{A}}^{-\frac{1}{2}} \widetilde{\mathcal{A}} \overline{\mathcal{A}}^{-\frac{1}{2}} \mathcal{Z}^{M-1} \textbf{w}^M), 
\label{eq_encoder}
\end{align}
where $\Phi^M(\cdot)$ represents a nonlinear activation function, e.g., $\tanh(\cdot)$ or $\text{ReLU}(\cdot)$; and $\overline{\mathcal{A}}^{-\frac{1}{2}} \widetilde{\mathcal{A}} \overline{\mathcal{A}}^{-\frac{1}{2}}$ is the symmetrically formulated adjacency matrix~\cite{zhu2020anomaly}. 

$\bullet$ \textbf{Decoder: }
The decoder is responsible for taking the lower-dimensional representation generated by the encoder, i.e., $\mathcal{Z}^M$ in~\eqref{eq_Z}, and mapping it back to the original $G({{\cal V}}, E, \mathcal{F})$. This can be viewed as the inverse operation of the encoder. The decoder aims to generate the original graph from its reduced representation. The output of the decoder is compared with the original input graph to evaluate a loss. 
The encoder and decoder are trained together to minimize the loss. 

A reconstructed adjacency matrix is generated at the decoder, which is defined as 
\begin{align}
\widehat{\mathcal{A}} = {\rm sigmoid} \left(\mathcal{Z}^M \left(\mathcal{Z}^{M}\right)^T\right).
\label{eq_decoder}
\end{align}
where the Sigmoid function is defined as ${\rm sigmoid} (x) = 1/(1+\exp(-x))$. The larger the inner product $(\mathcal{Z}^M \left(\mathcal{Z}^{M}\right)^T)$, the more likely the vertexes $j$ and $j^\prime$ are connected in the graph~\cite{pan2019learning}. 

The output of the decoder is the reconstructed adjacency matrix $\widehat{\mathcal{A}}$. A reconstruction loss function that measures the difference between $\mathcal{V}$ and $\widehat{\mathcal{A}}$ can be formulated as~\cite{qiu2022fast}
\begin{align}
\phi_{\rm loss} = \mathbb{E}_{f_G(\mathcal{Z}^{M-1}, G | \textbf{w}^M)} \Big[ \log~p(~\widehat{\mathcal{A}}~|~\mathcal{Z}^M~) \Big],
\label{eq_reconError}
\end{align}
where $p(~\widehat{\mathcal{A}}~|~\mathcal{Z}^M~)$ at the decoder indicates the correlation among the embedding vertexes, and is given by 
\begin{align}
p(~\widehat{\mathcal{A}}~|~\mathcal{Z}^M~) = \Pi_{j=1}^J \Pi_{j^\prime=1}^J p(~\widehat{\mathcal{A}}_{jj^\prime}~|~\mathcal{Z}^M_j, \mathcal{Z}^M_{j^\prime}~), 
\label{eq_pFunc1}
\end{align}
where 
\begin{align}
p(\widehat{\mathcal{A}}_{jj^\prime} = 1|\mathcal{Z}^M_j, \mathcal{Z}^M_{j^\prime}~) = {\rm sigmoid} \left(\mathcal{Z}^M_j\left(\mathcal{Z}_{j^\prime}^{M}\right)^T\right). 
\label{eq_pFunc2}
\end{align}

$\bullet$ \textbf{Malicious Model Generation: }
A graph signal processing module is designed to decompose the correlation features of the benign local models, and the data features substantiating the local models, as described earlier.
A Laplacian matrix~\cite{molitierno2016applications} is built based on the adjacency matrix of the benign models, i.e., $\mathcal{A}$, as given by 
\begin{align}
\mathcal{L} = diag(\mathcal{A}) - \mathcal{A}.
\label{eq_Laplacian}
\end{align}
By applying singular value decomposition (SVD)~\cite{lange2010singular} to $\mathcal{L}$, i.e., $\mathcal{L}=B\Sigma B^T$, we can obtain a complex unitary matrix $B \in \mathbb{R}^{J \times J}$, also known as graph Fourier transform (GFT) basis, that is used to transform graph data, e.g., $\mathcal{F}$, to its spectral-domain representation. $\Sigma$ is a diagonal matrix with the eigenvalues of $\mathcal{L}$ along its main diagonal. 

As a result, the attacker can obtain a matrix $S$ that contains the spectral-domain data features of all benign local models, by removing the correlations among the models and subsequently focusing on the data features substantiating the local models. $S$ is given by 
\begin{align}
S = B^{-1}\mathcal{F}.
\label{eq_S}
\end{align}

Likewise, the attacker can use the graph signal processing module to produce a Laplacian matrix based on the output of the GAE, as given by 
\begin{align}
\widehat{\mathcal{L}} = diag(\widehat{\mathcal{A}}) - \widehat{\mathcal{A}}.
\label{eq_maliciousLap}
\end{align}
The corresponding GFT basis, denoted by $\widehat{B}$, can be obtained by applying SVD to $\widehat{L}$. With reference to~\eqref{eq_S}, the malicious local model that follows $\mathcal{A}$ in the GAE can be determined by
\begin{align}
\widehat{\mathcal{F}} = \widehat{B} S, 
\label{eq_maliciousF}
\end{align}
where $\widehat{\mathcal{F}} \in \mathbb{R}^{J \times D}$. The vector $\pmb{\omega}^a(t)$ in $\widehat{\mathcal{F}}$ is selected as the malicious local model and uploaded by the attacker to the aggregator for global model aggregation in the $t$-th communication round. 

Since the attacker aims to generate the malicious local models to disorient FL, the proposed GAE is constructed and trained to maximize { $ L(\pmb{\omega}^a(t), \lambda(t))-\phi_{\rm loss}$.} 
As a consequence, the malicious local model $\pmb{\omega}^a(t)$  progressively and increasingly contaminates the FL training process with the increase in global model aggregations, i.e., $t=1,2,\cdots$.

\subsection{Training Algorithm of the Proposed GAE Model}
Algorithm~\ref{alg_flaph} summarizes the training process of the proposed GAE-based, data-agnostic, model poisoning attack model, which operates along with the FL training of the benign devices and the server. 
Specifically, in every FL communication round, i.e., the $t$-th round, the server broadcasts $\pmb{\omega}_g^a(t)$. The benign devices apply the LocalTraining\_start($\pmb{\omega}_g^a(t)$) function to train their local models $\pmb{\omega}_j(t),\, \forall j=1,\cdots,J$; see Steps~3 -- 5 in Algorithm~\ref{alg_flaph}.

On the other hand, the attacker overhears the local model $\pmb{\omega}_j(t),\,\forall j$ from the benign devices at the $t$-th FL communication round, and recall the global model $\pmb{\omega}_g^a(t-1)$ overheard from the server at the $(t-1)$-th round.
The GAE is trained to maximize the data-agnostic model poisoning attack problem in~\eqref{eq_opt} with $\mathcal{V}$ and $\mathcal{F}$. 
Specifically, the problem in \eqref{eq_opt} is transformed into a primal and a dual problem using the Lagrangian method. 
Given the dual variable $\lambda(t)$, $\pmb{\omega}^a(t)$ is optimized using the GAE;
see Steps 8 -- 10 in Algorithm~\ref{alg_flaph}. 
With the obtained ${\pmb{\omega}^a(t)}^*$, the sub-gradient descent method is taken to update $\lambda(t)$ by~\eqref{eq-sub-gradient}; see Step 11.
At the output of the GAE, the attacker achieves the optimal malicious local model, i.e., $\pmb{\omega}^a(t)$. Next, $\pmb{\omega}^a(t)$ is uploaded to the server for the next round of the FL training. As $\pmb{\omega}^a(t)$ is highly correlated with $\pmb{\omega}^a(t)$ from the benign user devices, the server is unable to identify the attacker.

\begin{algorithm}[t]
\caption{The proposed GAE-based, data-agnostic model poisoning attack against FL}
\label{alg_flaph}
\begin{algorithmic}[1]
\STATE{\textbf{1. Initialize}: $G(\cal{V}, E, \mathcal{F})$, $T_L$, $J$, $d_T$, $\pmb{\omega}_g^a(t)$, $\pmb{\omega}^a(t)$, and $\lambda(1) \geq 0$.}\\
{\% \textbf{Adversarial FL}:}
\FOR{round $t$ = 1, 2, 3, ...}
\FOR{Local iteration number $t_L = 1,\cdots, T_L$}
\STATE{All benign user devices train the benign local model $\pmb{\omega}_j(t)$, $j =1,\cdots, J$.}
\ENDFOR
\STATE{All benign user devices upload their benign local models $\pmb{\omega}_j(t)$, $j=1,\cdots,J$ to the server, and the attacker overhears the benign local models.}
\STATE{The attacker carries out the proposed GAE, i.e., {{\textbf{GAE}}({$\pmb{\omega}_j(t),\forall j,{\cal{F}}, \lambda(t)$}), and obtains $\pmb{\omega}^a(t)$, as follows.}}\\
\STATE{\hspace{4 mm} $\boldsymbol{\cdot}$ Calculate {the adjacency matrix $\mathcal{A} = \{\overline{\omega}_{j,j^\prime}\} \in$ \\ \hspace{5 mm}  $  \mathbb{R}^{J \times J}$ according to~\eqref{eq_innerProduct}, and input $\mathcal{A}$ and ${\cal{F}}$ \\ \hspace{5 mm}into the GAE. }}
\STATE{{\hspace{4 mm} $\boldsymbol{\cdot}$ Train the GAE to maximize the reconstruction \\ \hspace{4 mm}  loss {$ L(\pmb{\omega}^a(t), \lambda(t))-\phi_{\rm loss}$} to obtain $\widehat{\mathcal{A}}$.}}
\STATE{\hspace{4 mm} $\boldsymbol{\cdot}$ Obtain $S$ based on~\eqref{eq_Laplacian} and~\eqref{eq_S}, next obtain\\
\hspace{4 mm} $ \widehat{\mathcal{F}}$ based on~\eqref{eq_maliciousLap} and~\eqref{eq_maliciousF}, and then determine \\
\hspace{4 mm} $\pmb{\omega}^a(t)$ based on~$\widehat{\mathcal{F}}$.}
\STATE{Update {$\lambda(t)$}, according to \eqref{eq-sub-gradient}.}

\STATE{The attacker uploads the malicious local model $\pmb{\omega}^a(t)$ to the server.}
\STATE{The server aggregates all the local models to obtain the global model {under attack $\pmb{\omega}^a_g(t)$ by~\eqref{eq_glbAttacks}, and broadcasts $\pmb{\omega}^a_g(t)$}.}
\STATE{All benign user devices update their local models with the global model, i.e., {$\pmb{\omega}_j(t) \leftarrow \pmb{\omega}^a_g(t),\,\forall j$}.}
\ENDFOR
\end{algorithmic}
\end{algorithm}

\subsection{Convergence Analysis of FL under Attack}
We derive the convergence upper bound for the FL under the proposed, data-agnostic, model poisoning attack. The following assumptions are made before the analysis, as typically considered in the literature~\cite{wei2020federated,truex2020ldp,zhao2020local}.
\begin{assumption}
\label{assumption}
	$\forall m \in {\cal M}$,
	\begin{enumerate}
	    
	\item The gradient of $F_j(\pmb \omega_j)$ is
         $L$-Lipschitz continuous~\cite{o2006metric}, that is, 
         $\left\|\nabla F_j ({\pmb \omega_j}(t+1))-\nabla F_j ({\pmb \omega_j}(t)) \right\| \leq L \left\| {\pmb \omega_j}(t+1) - {\pmb \omega_j}(t)\right\|,\,\forall {\pmb \omega_j}(t+1),{\pmb \omega_j}(t) $, 
         with $L$ being a constant depending on the loss function so that the gradient of the global loss function is also $L$-Lipschitz continuous; 
                 
    \item $F_j(\pmb{\omega}_j)$ is $L_c$-Lipschitz continuous; in other words, $\left| F_j({\pmb{\omega}_j}) - F_j({\pmb{\omega}_j'}) \right| \leq L_c \left\| {\pmb{\omega}_j} - {\pmb{\omega}_j'}\right\|,\,\forall {\pmb{\omega}_j},{\pmb{\omega}_j'}$;
    
	\item The learning rate is $\eta \leq \frac{1}{L}$; 
		
	\item At device $j$, the expected squared norm of the stochastic gradients is uniformly bounded by $\mathbb{E}\left\| \nabla F_j ({\pmb{\omega}}_j(t)) \right\|^2  \leq \kappa \|\nabla F(\pmb{\omega}_{g}(t))\|^2, \forall j,\kappa \geq 0$;
		
	\item With $\rho\geq 0$, $F_j(\pmb \omega)$ fulfills the Polyak-Lojasiewicz requirement~\cite{karimi2016linear}, indicating that $F(\pmb \omega_g)-F({\pmb \omega}_g^*) \leq \frac{1}{2\rho} \left\| \nabla F(\pmb \omega_g) \right\|^2 $, where ${\pmb \omega^*_g}=\arg\min_{\pmb \omega_g}F(\pmb \omega_g)$;
	
    \item $F({\pmb \omega}_g(0) - F({\pmb \omega}_g^*) = \Theta$, where $\Theta$ is a constant.
	\end{enumerate}
\end{assumption}
Under Assumption~\ref{assumption}, we develop the following theorem that provides the convergence bound of the gap between ${\pmb \omega}_g(t), \forall t$ and ${\pmb \omega}_g^*$. 
\begin{theorem}\label{theo_convergence bound}
	At the $t$-th communication round, the convergence upper bound of the attacked FL is obtained as
	\begin{equation}\label{eq_convergnece_bound 0}
	\begin{aligned}
	 F({\pmb \omega}^a_{g}(t) )& - F({\pmb \omega}_g^*) \leq \Theta  \zeta^t + \frac{1-\zeta^t}{1-\zeta} \cdot \frac{\rho \eta D D_a}{(D-D_a)^2} F^{\max},
	\end{aligned}
	\end{equation}	
where $\zeta = 1 - \frac{\rho \eta D^2}{(D-D_a)^2}$, and $F^{\max}$ is a maximum value of $F (\pmb{\omega}^a(t))$ due to the constraint in~\eqref{eq_opt_obj} and~\eqref{eq_const_dist}, i.e., $ F (\pmb{\omega}^a(t)) \leq F^{\max}$. 
\end{theorem}
\begin{proof}
	See Appendix~\ref{appendix_convergence bound}.
\end{proof}

%

As stated in Theorem~\ref{theo_convergence bound}, despite the attack launched by the attacker, the global model of FL can still converge, but to an inferior global model. Specifically, as $t \to \infty$, the optimality gap would stabilize at $\frac{2D_a L_c d_T}{D-D_a}$, which cannot be further reduced by training.


\begin{figure}
\begin{center}
\includegraphics[width=3in]{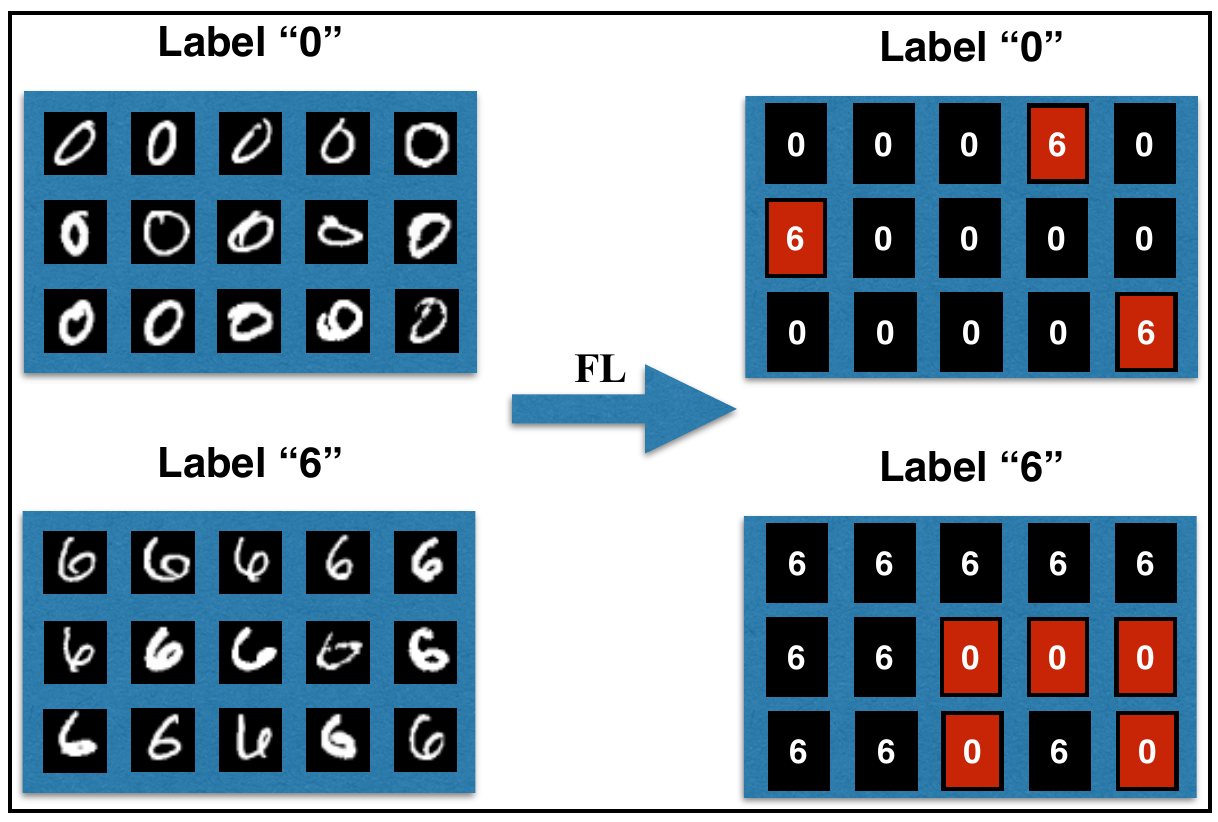}\\
(a) MNIST\\
\includegraphics[width=3in]{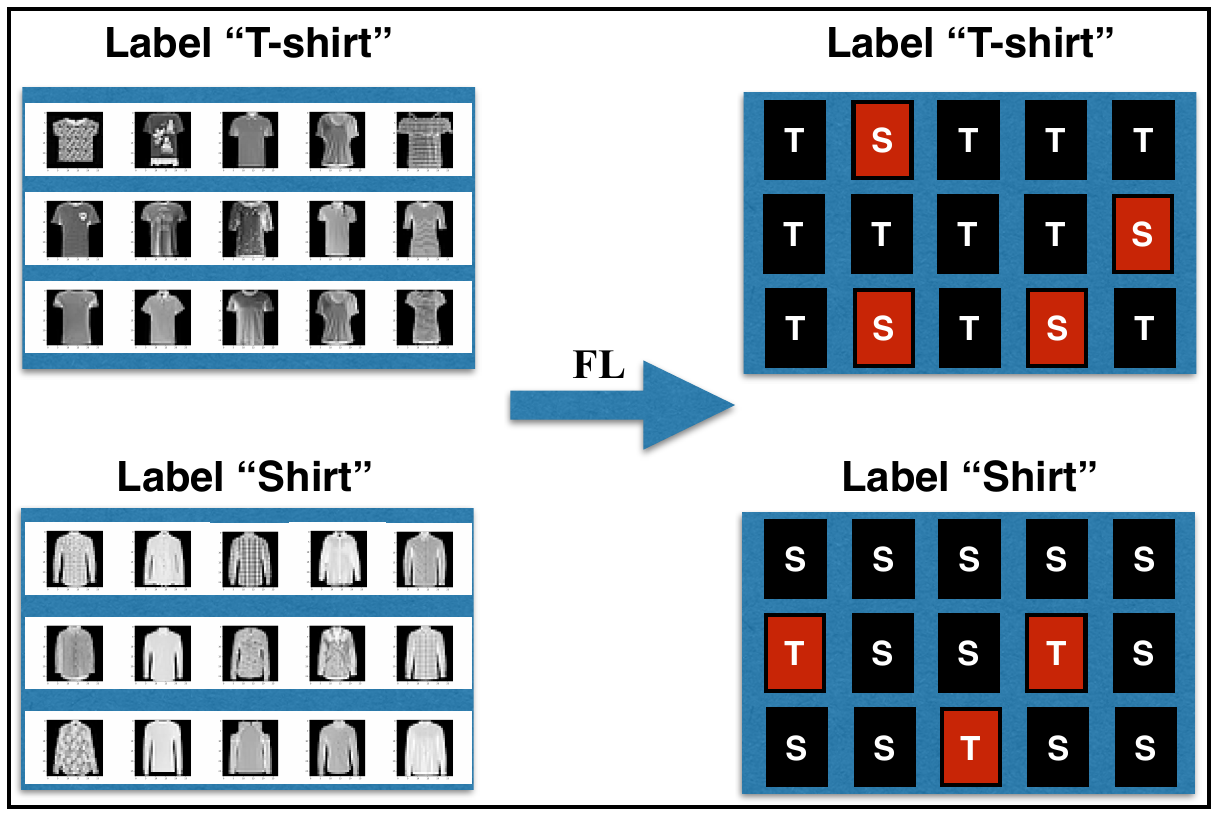}\\
(b) fashionMNIST\\
\includegraphics[width=3in]{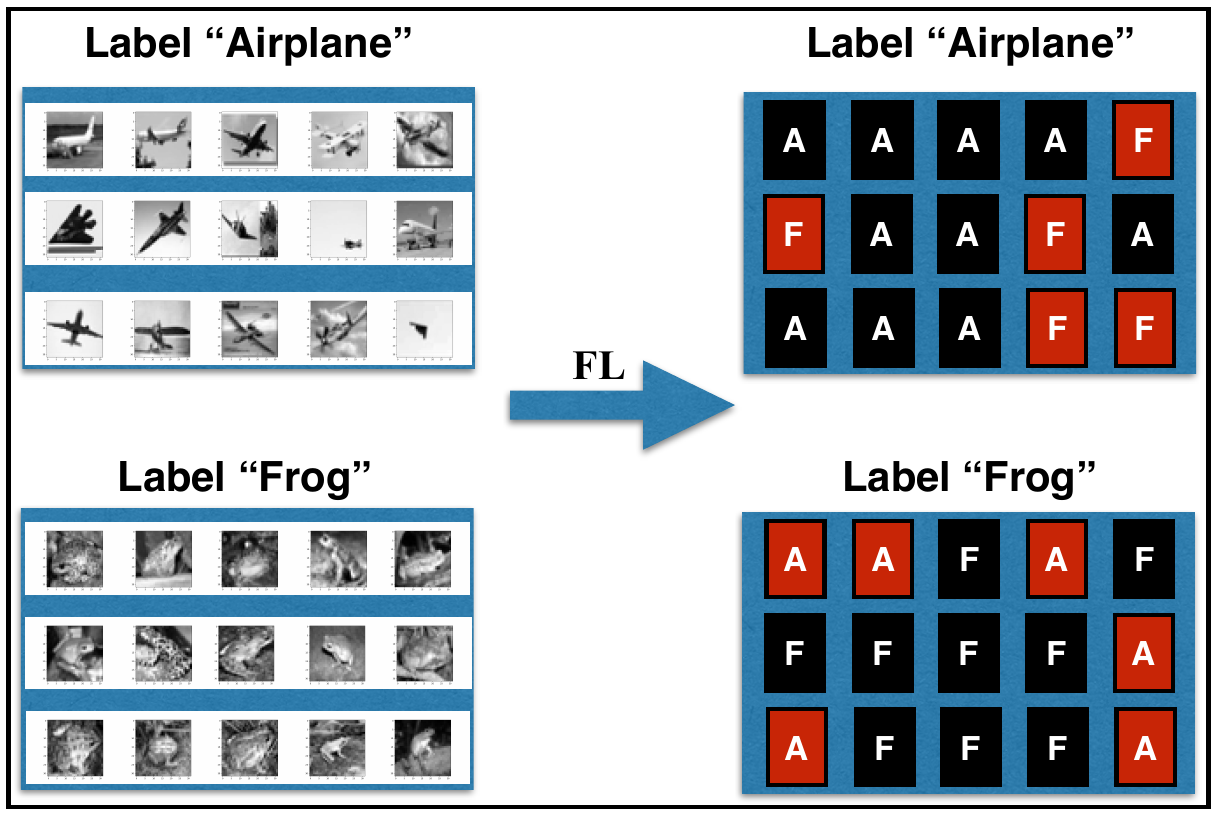} \\
(c) CIFAR-10
\end{center}
\caption{An illustration of the local model of a user device trained to classify images.}
\label{fig_datasets}
\end{figure}

\section{Performance Evaluation}
\label{sec_evaluation}
In this section, we present the implementation of the proposed GAE-based, data-agnostic, model poisoning attack in PyTorch. We evaluate the testing accuracy of the local and global models of FL under attack, using the MNIST~\cite{deng2012mnist}, fashionMNIST~\cite{xiao2017fashion}, and  CIFAR-10 datasets. 
We also report the detection rate of the attack, where the detection is based on the Euclidean distances of the malicious local models and the benign local models to the global models, as typically done in the latest literature, e.g.,~\cite{nguyen2022flame,cao2019understanding}. 

Moreover, we compare the proposed attack with the existent data-agnostic model poisoning (MP) attack that produces malicious local models by mimicking other benign devices' training samples to degrade the learning accuracy. As discussed earlier in Section~\ref{sec_GAE}, our GAE-based attack represents a novel type of attack, which only depends on the benign local models overheard and the global models, has no access to any of the training data, and attempts to compromise FL training processes.  
Few existing techniques can produce malicious models in such a way, i.e., based solely on the overheard benign models, as most existing techniques would require the knowledge of (part of) the dataset used in the FL training processes, e.g., for a different purpose, such as inserting a backdoor~\cite{nguyen2022flame} or injecting malicious traffic into the benign training dataset~\cite{nguyen2020poisoning}. The MP attack mechanism considered for comparisons with our proposed new attack has been implemented in several existing studies, e.g.,~\cite{cao2022mpaf} and~\cite{zhang2020poisongan}, in which the attacker manipulates the training process by injecting a fake device and sending fake local models to the server. 


\begin{figure*}[htb]
\begin{center}
\begin{tabular}{cc}
\includegraphics[width=3in]{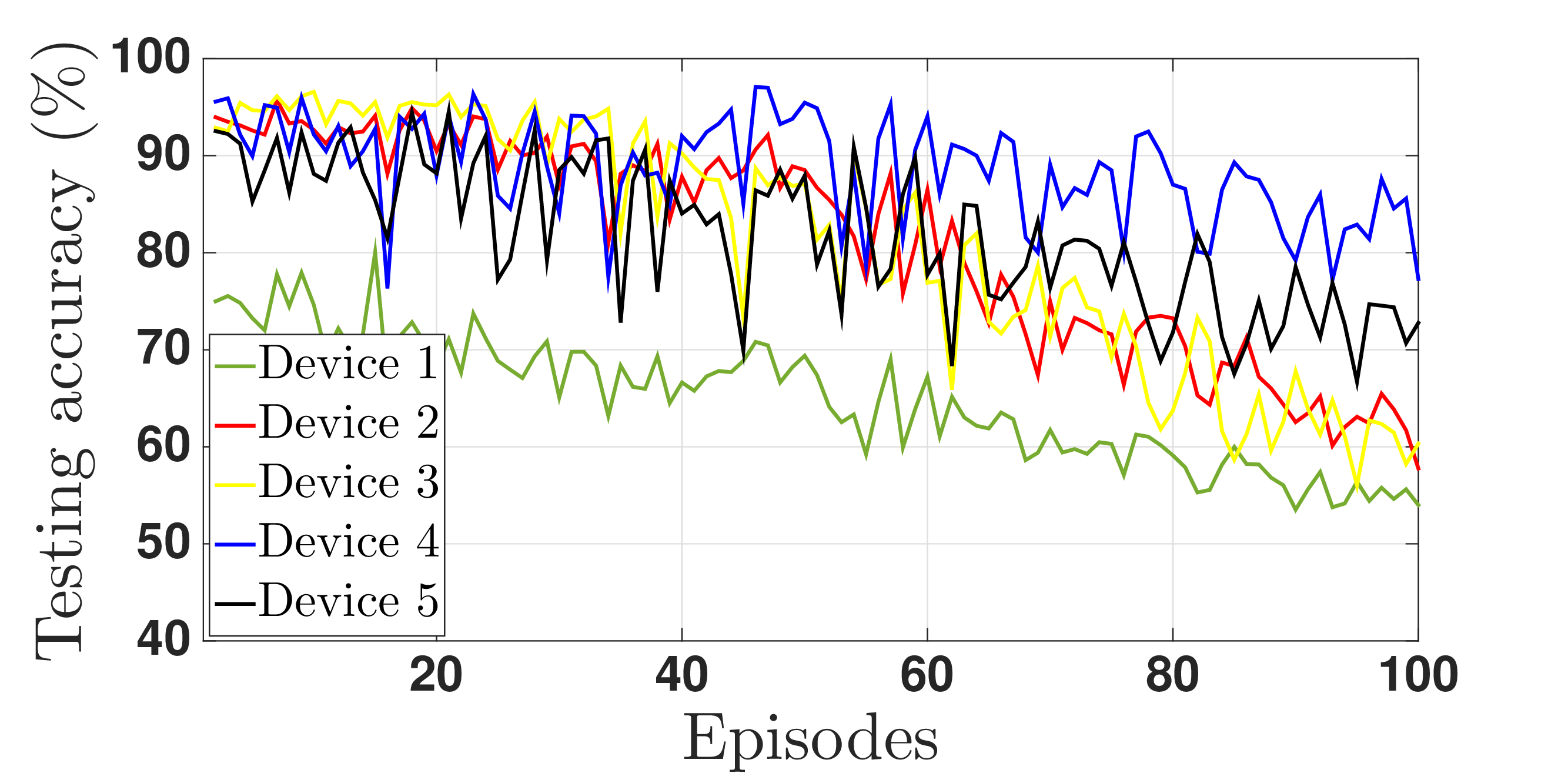} & \includegraphics[width=3in]{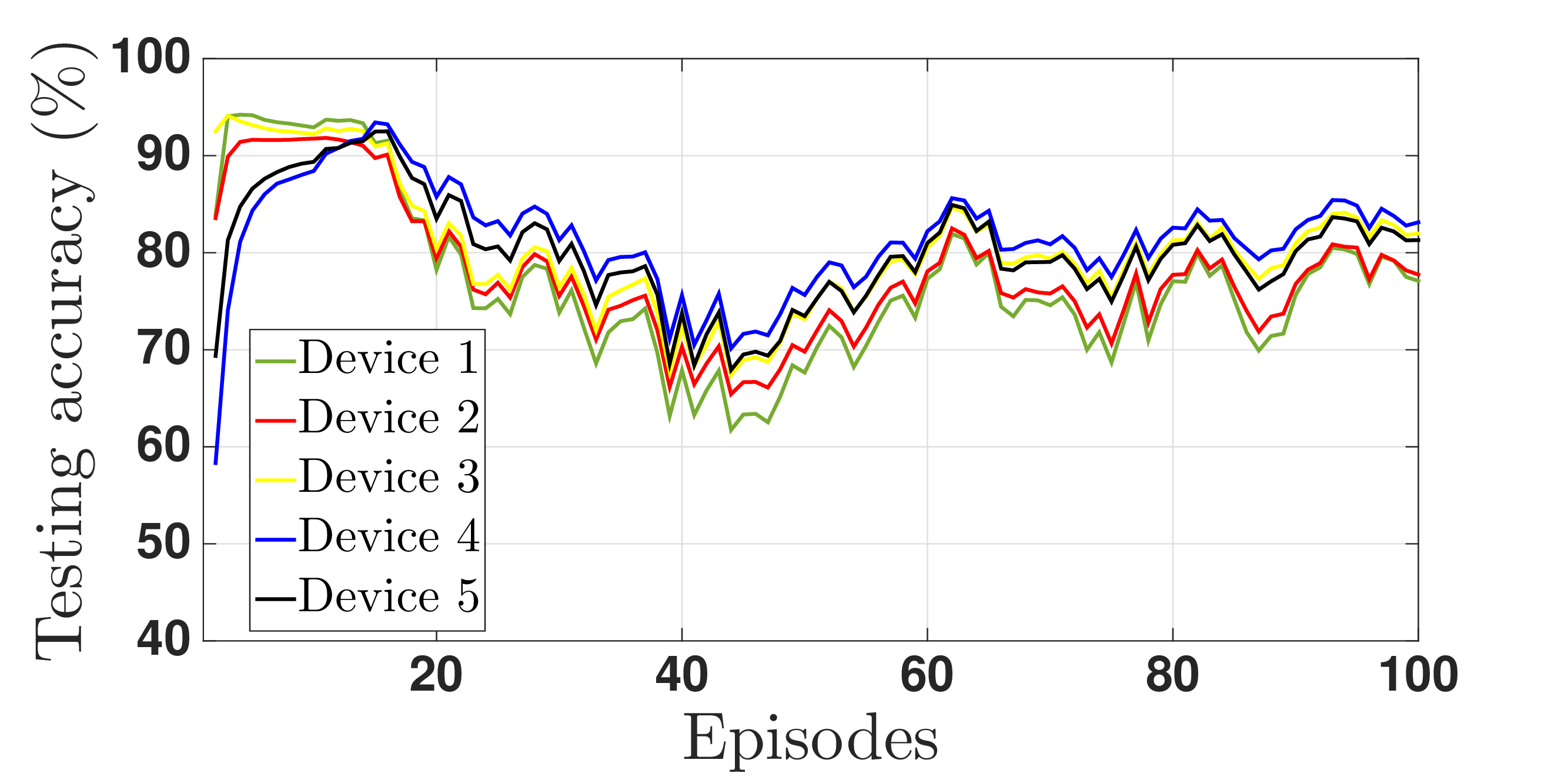}
\\ (a) The GAE-based attack with MNIST. & (b) The MP attack with MNIST.
\\ \includegraphics[width=3in]{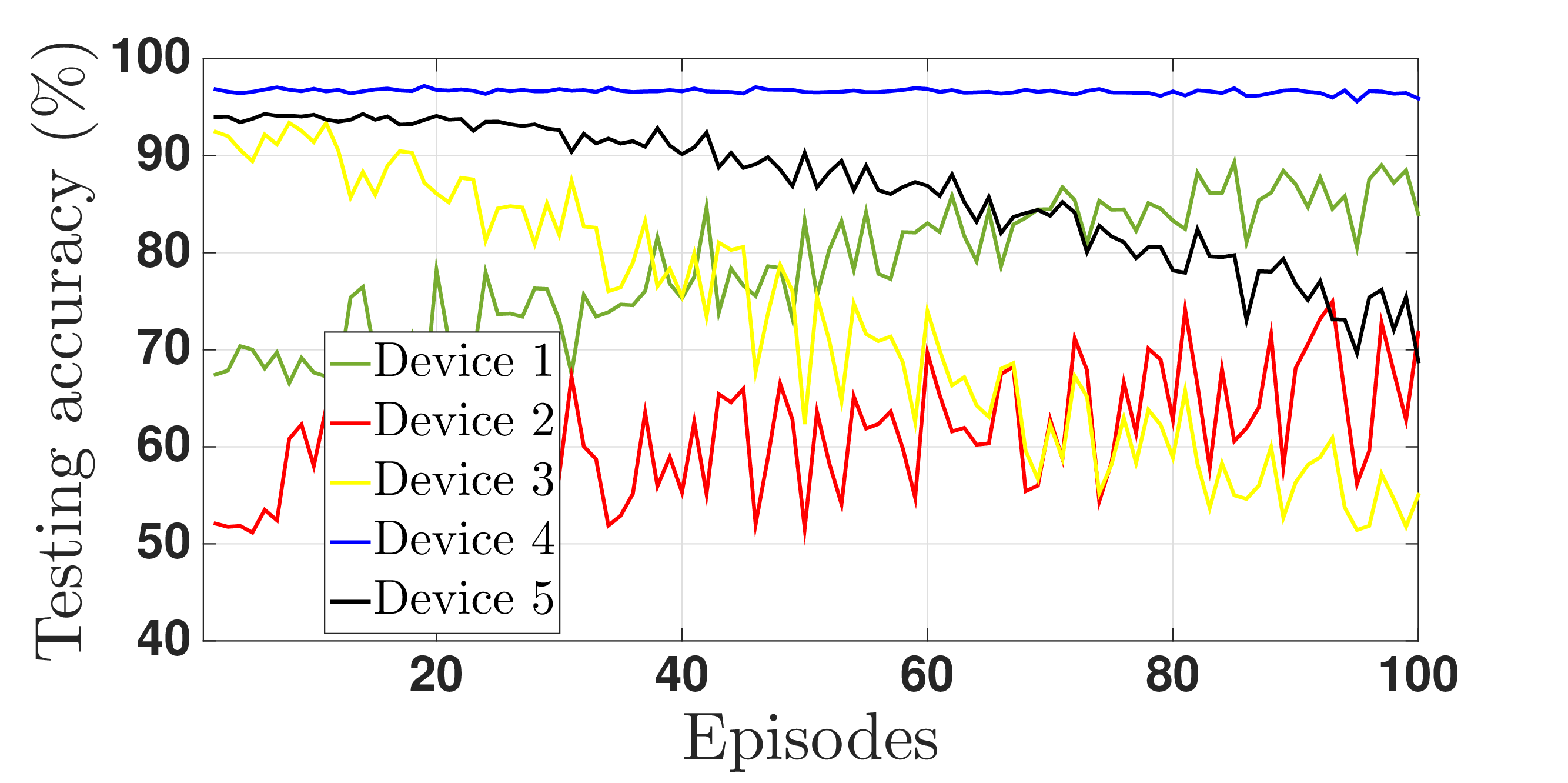} & \includegraphics[width=3in]{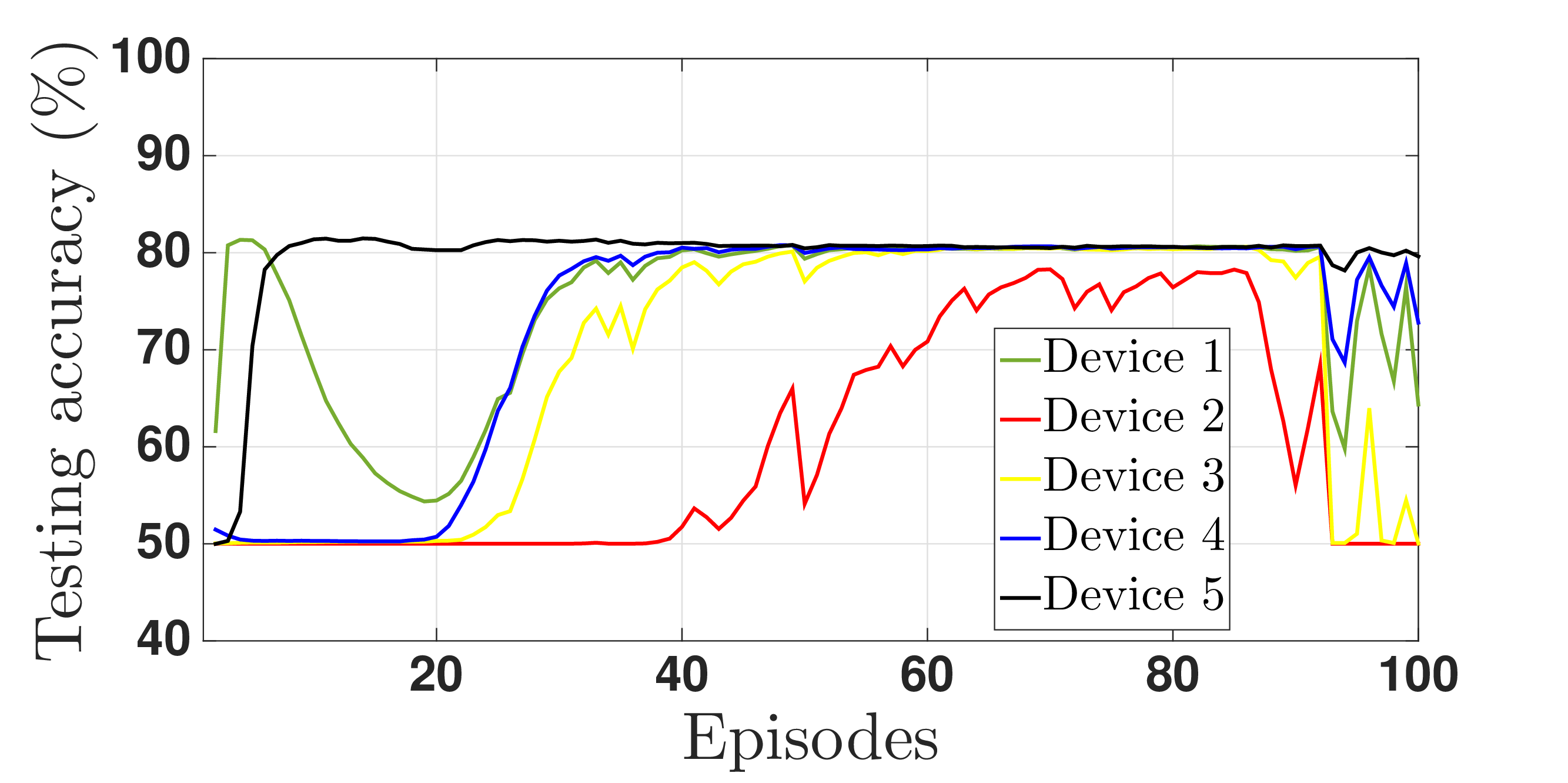} 
\\ (c) The GAE-based attack with fashionMNIST. & (d) The MP attack with fashionMNIST.
\\ \includegraphics[width=3in]{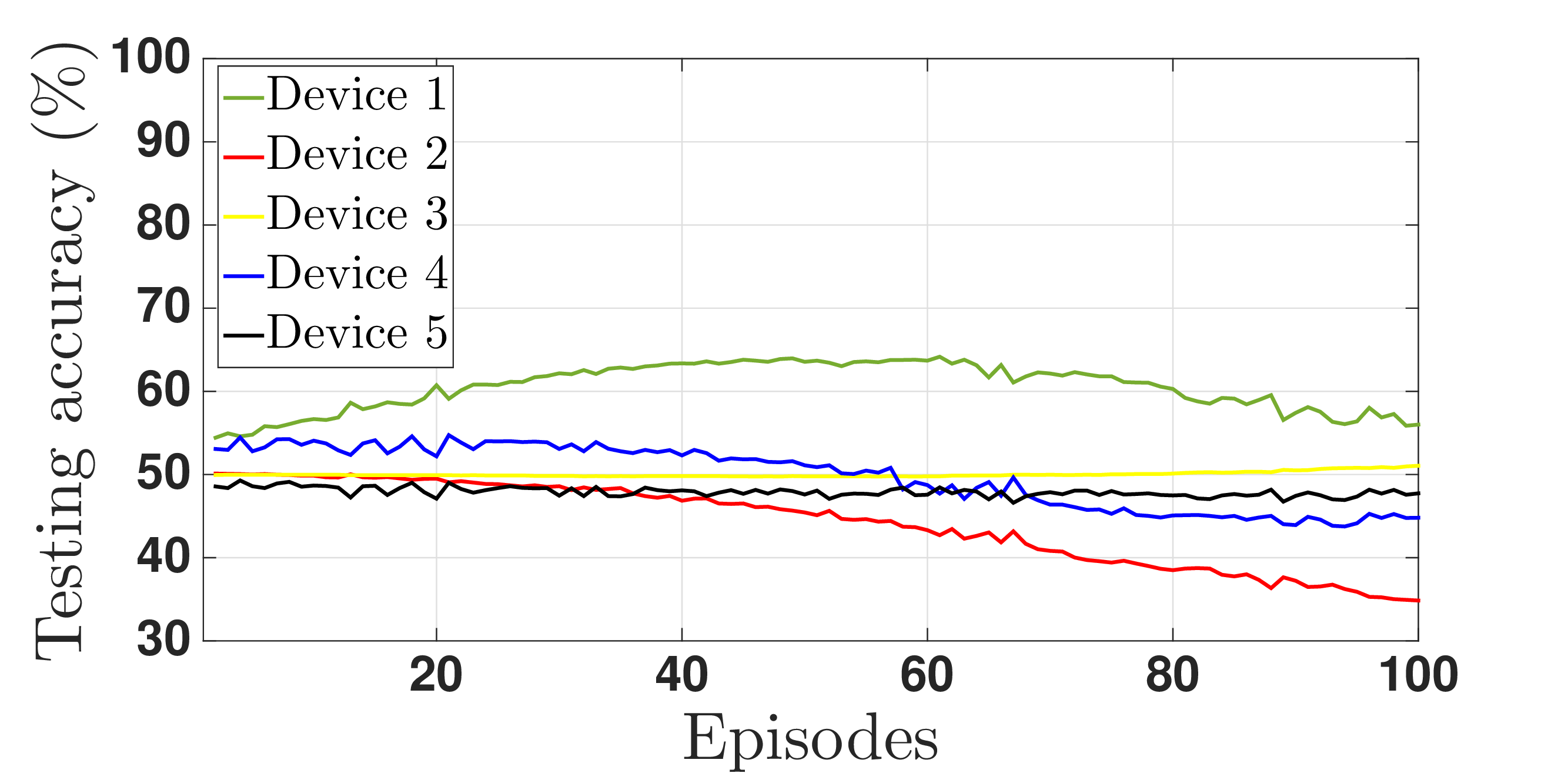} & \includegraphics[width=3in]{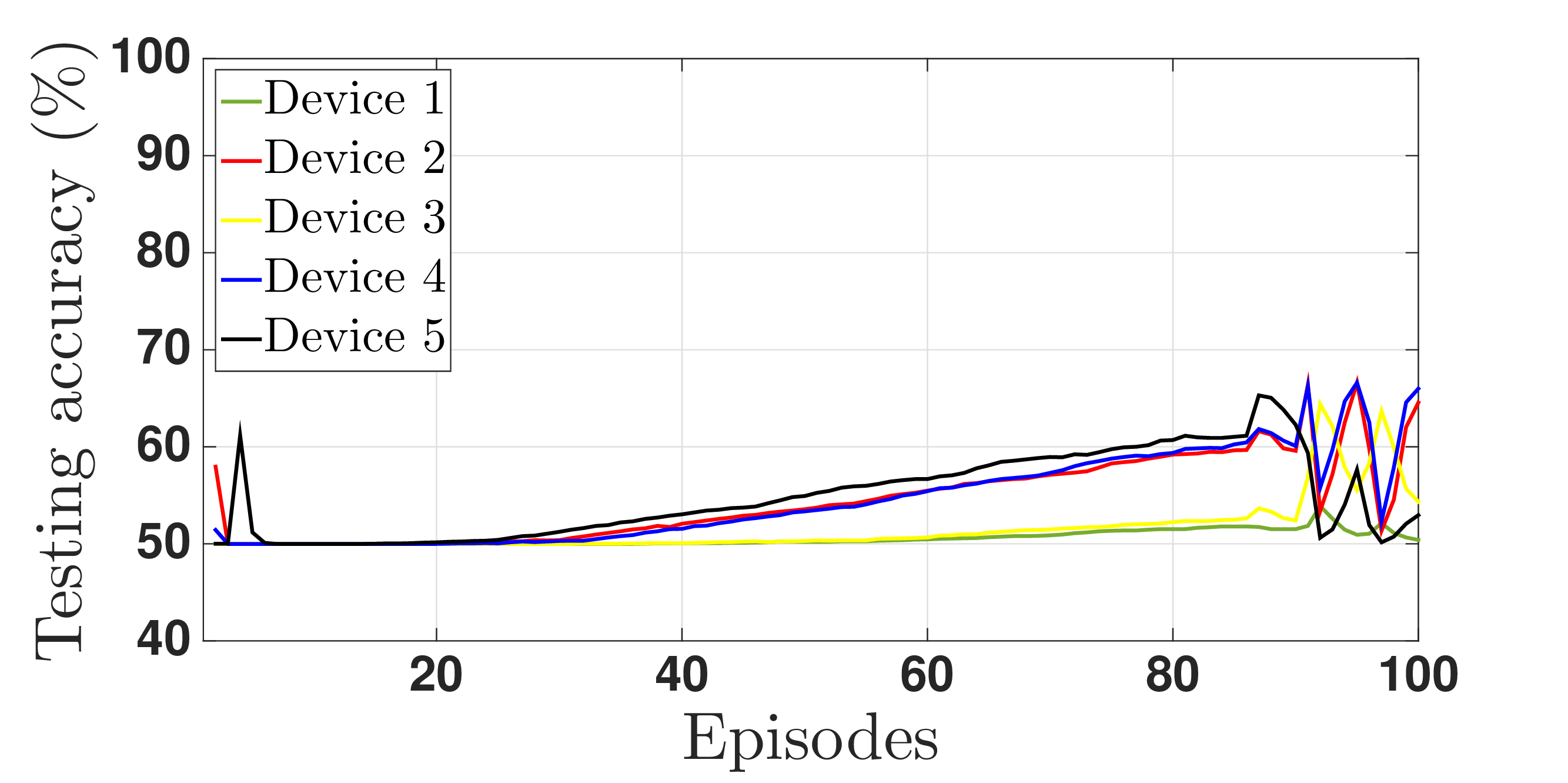} 
\\ (e) The GAE-based attack with CIFAR-10. & (f) The MP attack with CIFAR-10.
\end{tabular}
\end{center}
\caption{Given 100 FL communication rounds and five benign user devices, we compare the local model testing accuracy under the GAE-based attack and the existing MP attack on the MNIST, fashionMNIST, and CIFAR-10 datasets.}
\label{fig_accRounds}
\end{figure*}

\begin{figure}[htb]
\centering
\includegraphics[width=3.5in]{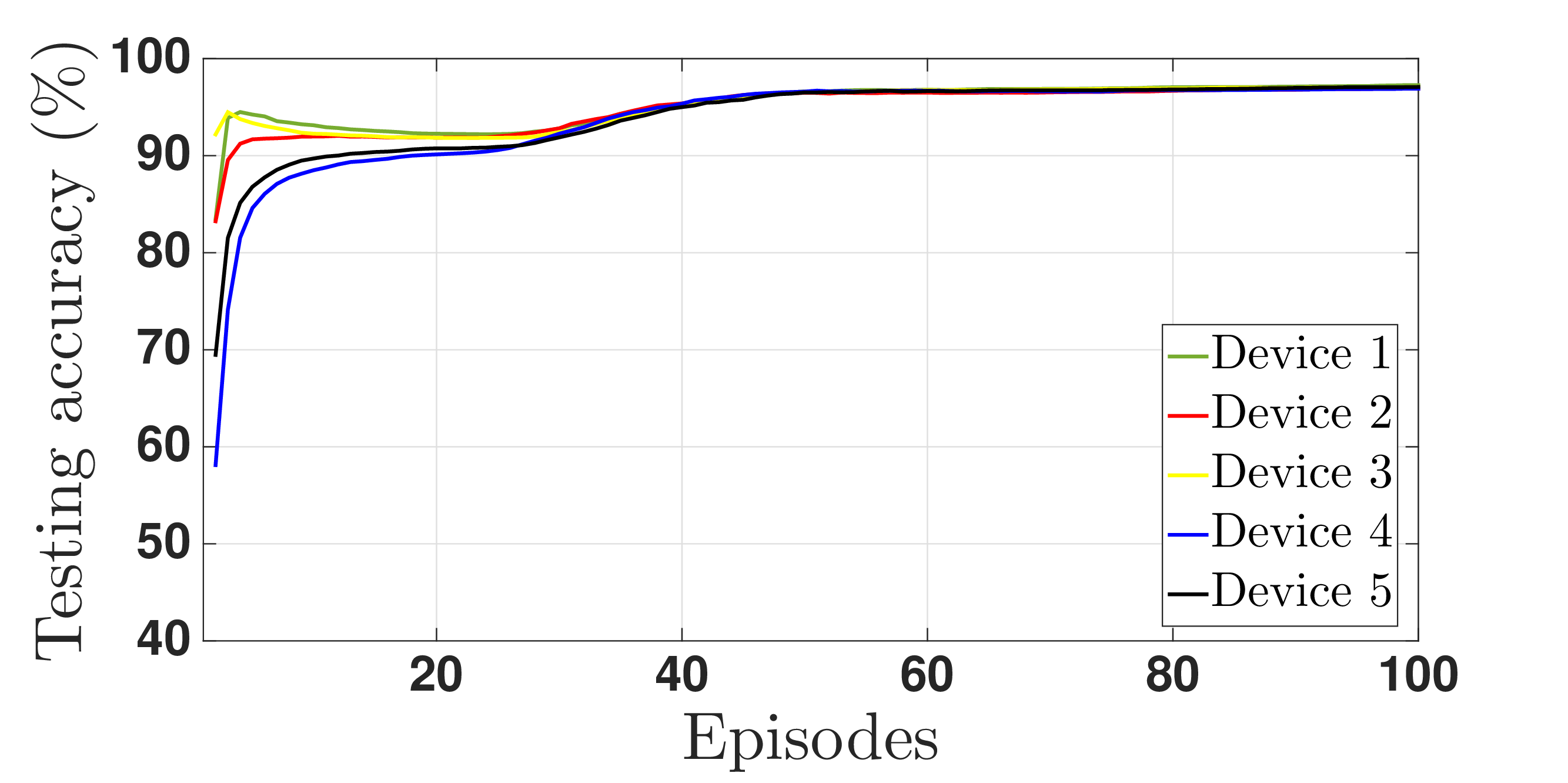}
\caption{The global model accuracy with no attack.}
\label{fig_accFL}
\end{figure}

\subsection{Implementation with PyTorch}
The number of benign devices is set to $J =5, 10, 15, 20, 25$. The number of iterations per communication round is set to $T_L = 10$. The maximum number of communication rounds is $T_{FL} =200$. By default, one attacker is considered, unless otherwise specified.

We implement the proposed GAE-based attack against the FL on an SVM model using PyTorch 1.12.1, Python 3.9.12 on a Linux workstation with an Intel(R) Core(TM) i7-9700K CPU@3.60GHz (8 cores) and 16 GB of DDR4 memory@2400 MHz.

The experiments are conducted on three datasets:
\begin{itemize}
	\item The standard MNIST dataset comprises 60,000 training examples and 10,000 testing examples, which are grayscale images of handwritten digits from 1 to 10; 

	\item The fashionMNIST dataset, which contains Zalando's article images (i.e., $28 \times 28$ grayscale images) in ten classes, including 60,000 examples for training and 10,000 examples for testing;

	\item The CIFAR-10 dataset, which contains 60,000 images with the size of $32 \times 32$ in ten classes (6,000 per class), 50,000 for training and 10,000 for testing.
\end{itemize}

At each user device, we use a standard quadratic optimization algorithm to train the SVM models based on the three datasets, namely, the standard MNIST, fashionMNIST, and CIFAR-10. 
The loss function used for training the SVM models is $F_j(\pmb{\omega}_j(t)) = \frac{1}{2}\left\|\pmb{\omega}_j(t) \right\|_2^2 + \frac{1}{D_j}\sum_{i=1}^{D_j}\max\left\lbrace 0, 1 - y_j^i ({\beta}_j + {\pmb{\omega}^T_j(t)} x_j^i) \right\rbrace$, where ${\beta}_j$ is a feature parameter based on ${\pmb{\omega}_j(t)}$~\cite{yang2019scheduling}. The global model $\pmb{\omega}^a_g(t)$, which is trained at the server according to~\eqref{eq_glbAttacks}, is broadcast to all user devices for the training of $\pmb{\omega}_j(t+1)$ in the next, $(t+1)$-th communication round. 
We note that, regardless of the specific model architecture (NNs or SVMs) employed for model training at benign user devices, the fundamental premise of the data-agnostic, model poisoning attack remains valid. Specifically, it involves the creation of malicious local models, which, when aggregated, compromise the global model by increasing the FL training loss. It also involves a new adversarial GAE crafted to generate these malicious local models based on the benign local models overheard, and captures the correlation features intrinsic to the benign local models and the global model, where the benign local models can be either NN or SVM models.

Fig.~\ref{fig_datasets} illustrates an example of label classification with the three datasets. In the MNIST dataset, three images labeled as ``0" are misclassified as ``6" while five images labeled as ``6" are misclassified as ``0", resulting in an FL accuracy of 73.3\%. Similarly, the FL accuracy with the fashionMNIST and CIFAR-10 datasets is 76.7\% and 63.3\%, respectively. The FL is designed to improve classification accuracy, while the proposed GAE-based attack aims to reduce accuracy and cause label misclassification. The GAE encoder is a two-layer GCN network (i.e., $M=2$) with a dropout layer to prevent overfitting. The GAE decoder is an inner product. We use the Adam optimizer with a learning rate of 0.01 to optimize the network. For all datasets, we use the same encoder, decoder and SVM models.

\subsection{Performance Analysis}

\begin{figure}[htb]
\centering
\includegraphics[width=3.5in]{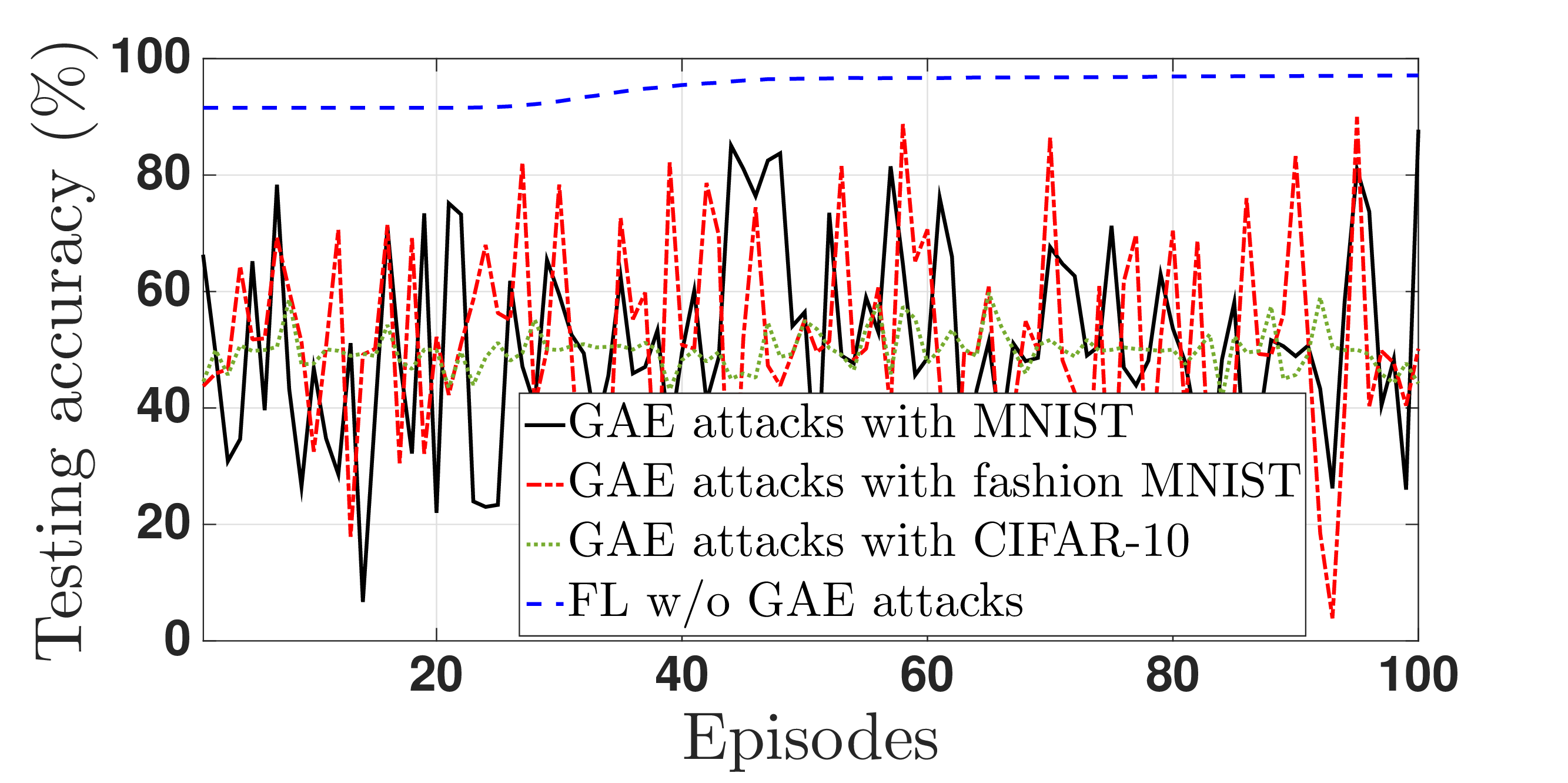}
\caption{The global model accuracy under the new attack.}
\label{fig_accGlobal}
\end{figure}

\subsubsection{FL Accuracy under Attack}
Fig.~\ref{fig_accRounds} plots the accuracy of the local models under the proposed GAE-based, data-agnostic, model poisoning attack on the MNIST, fashionMNIST, and CIFAR-10 datasets, where there are five benign devices (i.e., $J=5$) and 100 communication rounds for the FL. We compare the proposed attack with a model poisoning (MP) attack~\cite{nguyen2020poisoning}, in which the attacker manipulates the training process by injecting a fake device and sending fake local models to the server. 
Since the MP attack in~\cite{nguyen2020poisoning} shares the same objective as our proposed data-agnostic model poisoning approach, i.e., reducing the accuracy of FL, a comparison with this reference showcases the efficacy of our proposed method in the context of prevailing model poisoning attacks. 
For comparison purposes, Fig.~\ref{fig_accFL} plots the accuracy of the benign local models without any attacks. In this scenario, the accuracy of the user device can be improved efficiently by FL and rapidly converge to 96\%.

In Figs.~\ref{fig_accRounds}(a) and \ref{fig_accRounds}(b), we show that when using the MNIST dataset, the accuracy of all five devices under the proposed GAE-based attack gradually decreases and fluctuates dramatically. The performance of devices 1 and 2 drops from 75\% to 55\% and from 92\% to 59\%, respectively. The accuracy of the model drops from 91\% to 80\% when exposed to the MP attack in which the performance of the five devices follows a similar pattern. This is because the new GAE-based attack reconstructs the adversarial adjacency matrix according to the individual features of the user devices. As a result, the attacker falsifies the local models to maximize the FL loss; see~\eqref{eq_dual_equation}.

As shown in Figs.~\ref{fig_accRounds}(c) and~\ref{fig_accRounds}(d), the accuracy of device 3 and device 5 drops significantly by 37\% and 24\%, respectively, when using the fashionMNIST dataset and the proposed GAE-based attack. However, while the accuracy of devices 1 and 2 may slightly increase, their convergence rates are greatly slowed in comparison to Fig.~\ref{fig_accFL}. Additionally, the accuracy under the MP attack varies between 50\% and 80\%, with a minimal decrease in accuracy observed.

In Fig.~\ref{fig_accRounds}(e), it can be seen that the proposed GAE-based attack with the CIFAR-10 dataset greatly hinders the performance of FL, as the accuracy of all four user devices falls below 50\%. In contrast, the accuracy of all five devices under the MP attack is above 50\%, as shown in Fig.~\ref{fig_accRounds}(f).
Furthermore, it can be observed that the accuracy with the CIFAR-10 dataset is generally lower than the performance with the MNIST and fashionMNIST datasets. This is because the CIFAR-10 dataset contains a more diverse set of images, which makes it more challenging to differentiate and label, leading to lower overall accuracy.

Fig.~\ref{fig_accGlobal} illustrates the accuracy of the global model at the model aggregator. It can be observed that the proposed GAE-based attack hinders the training convergence when compared to the performance without the attack. As a result of the infection of the local FL model, the accuracy with the MNIST, fashionMNIST, or CIFAR-10 dataset fluctuates around 82\%, 81\%, or 25\%, respectively.

\begin{figure*}[htb]
\begin{center}
\begin{tabular}{cc}
\includegraphics[width=3in]{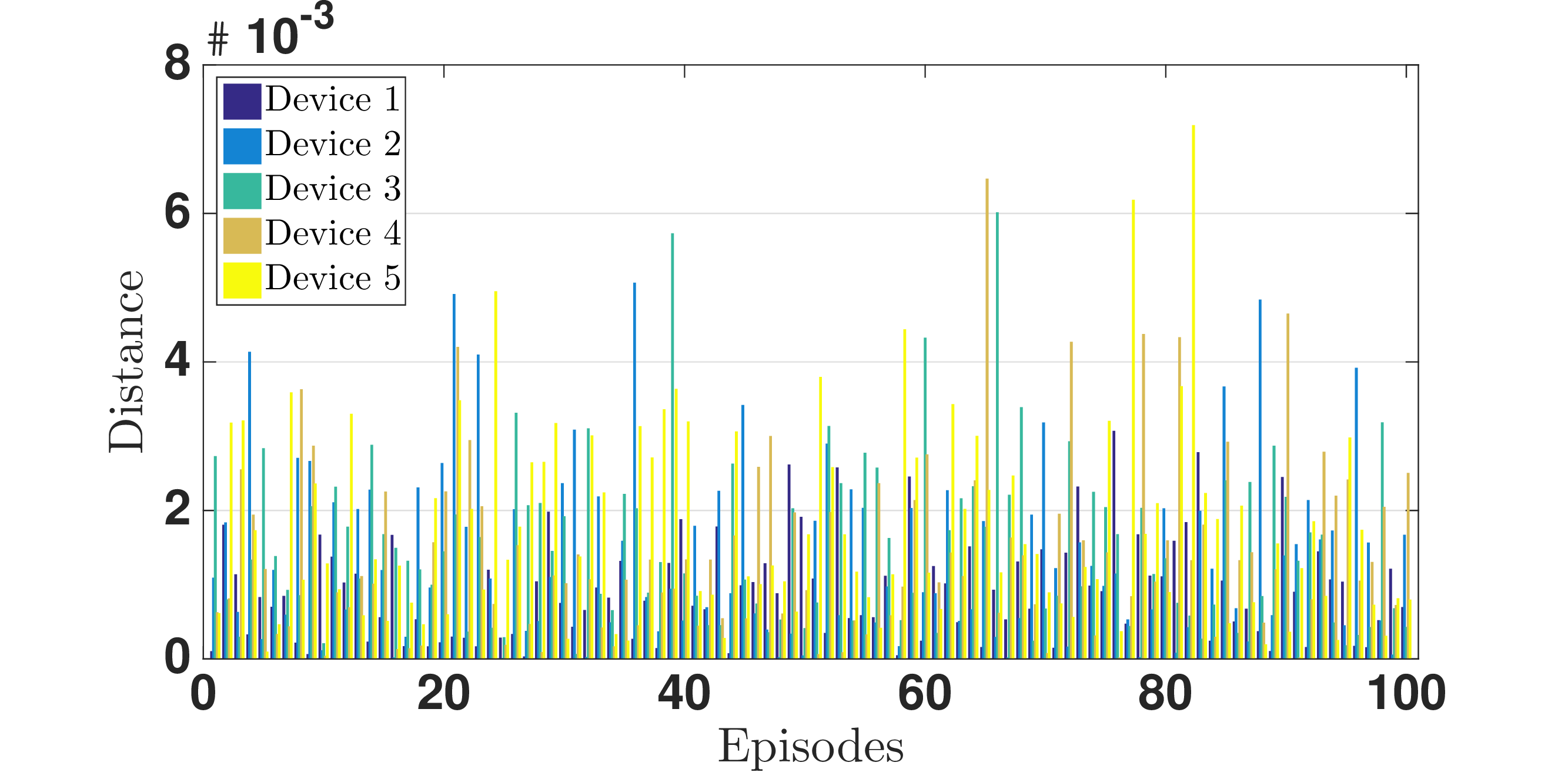} & \includegraphics[width=3in]{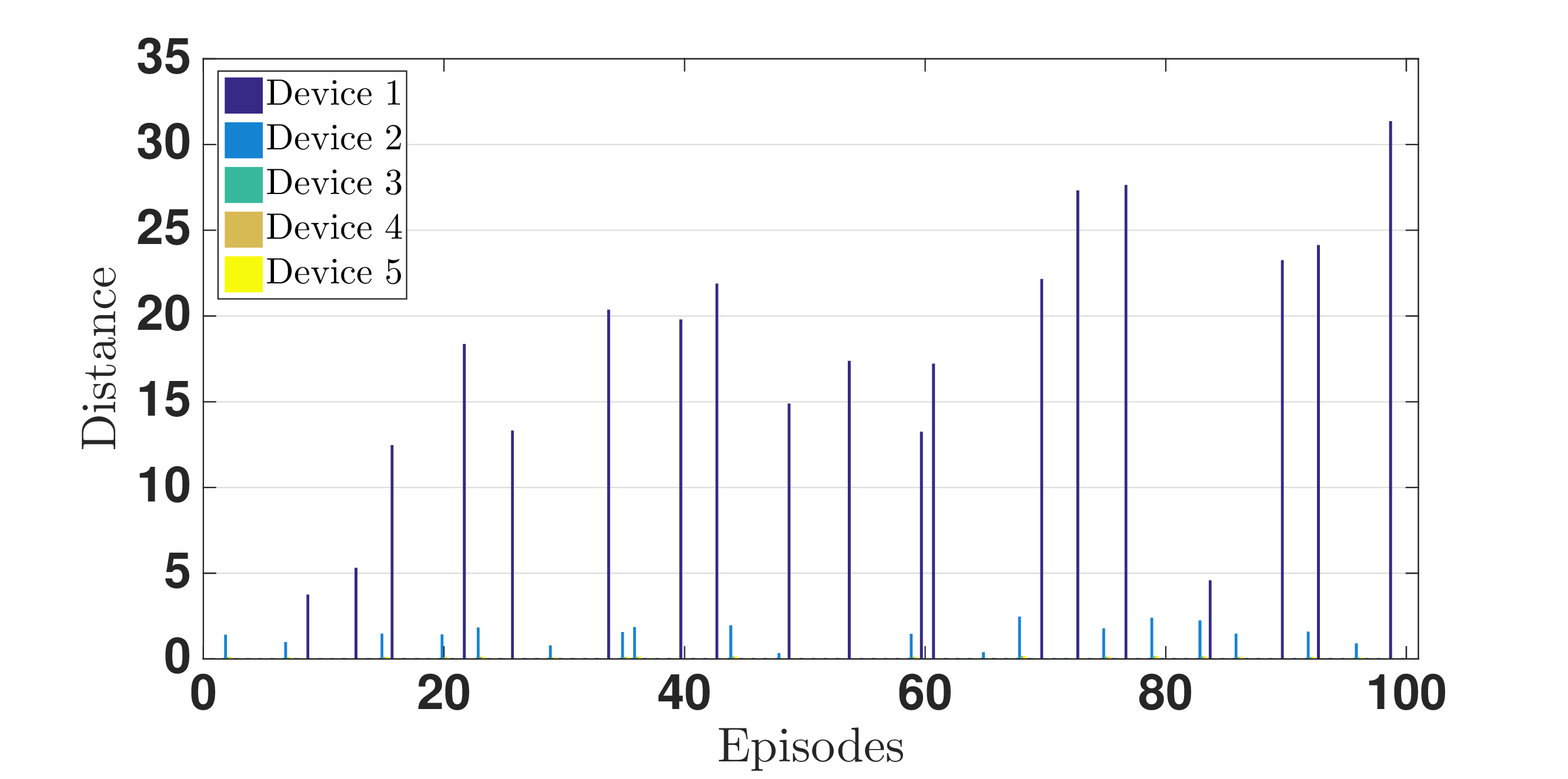}
\\ (a) The GAE-based attack with MNIST. & (b) The MP attack with MNIST.
\\ \includegraphics[width=3in]{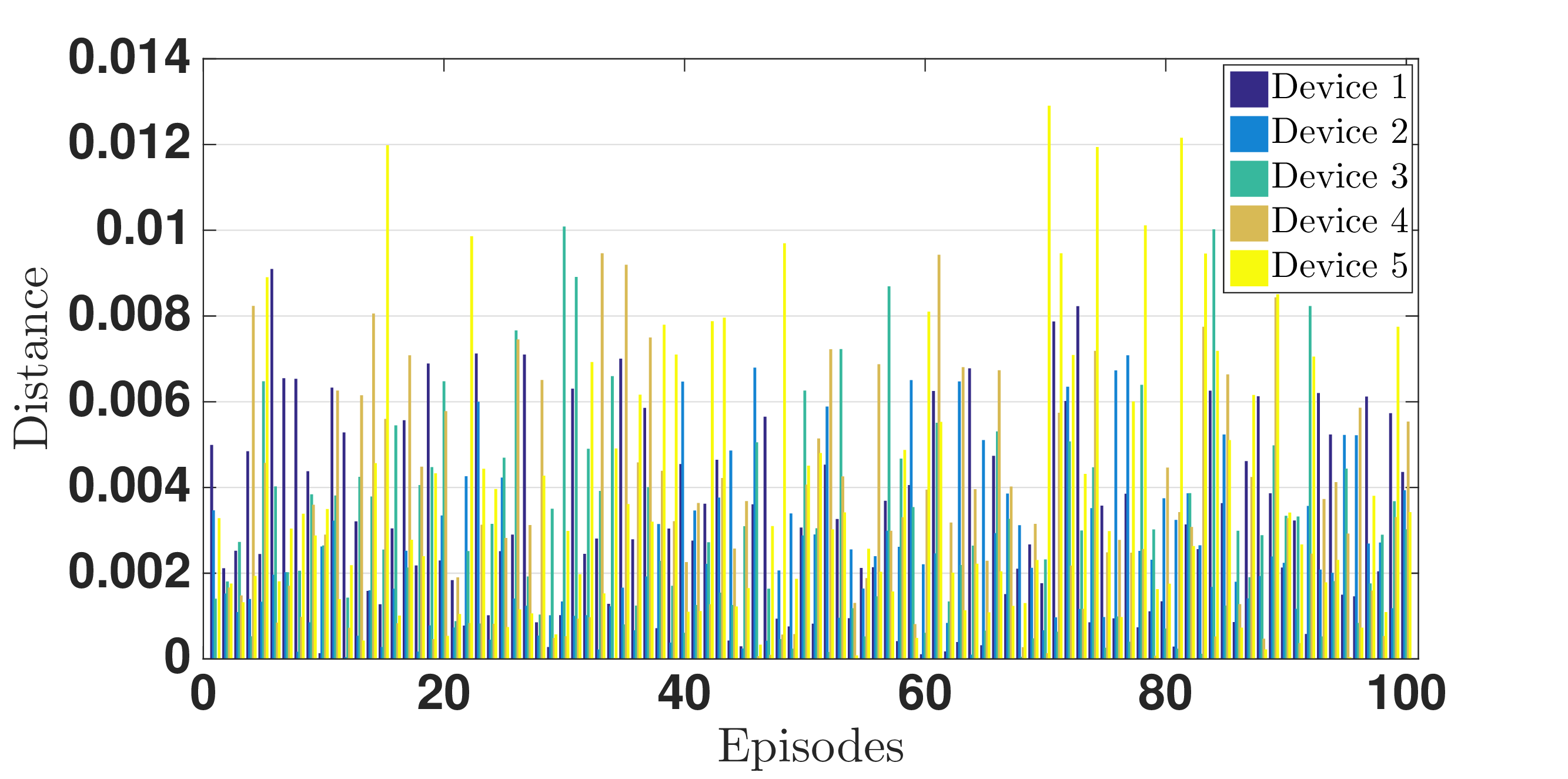} & \includegraphics[width=3in]{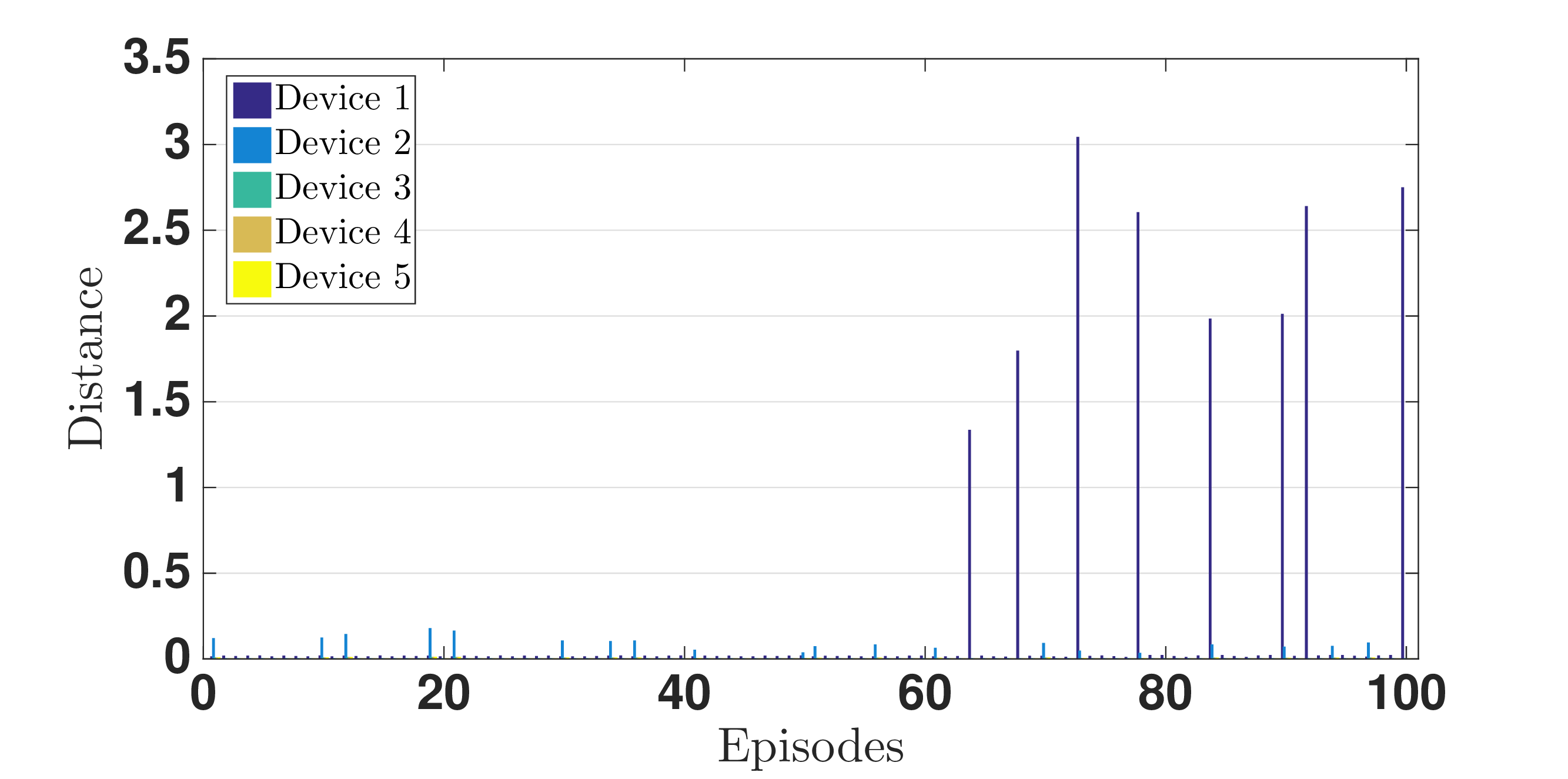} 
\\ (c) The GAE-based attack with fashionMNIST. & (d) The MP attack with fashionMNIST.
\\ \includegraphics[width=3in]{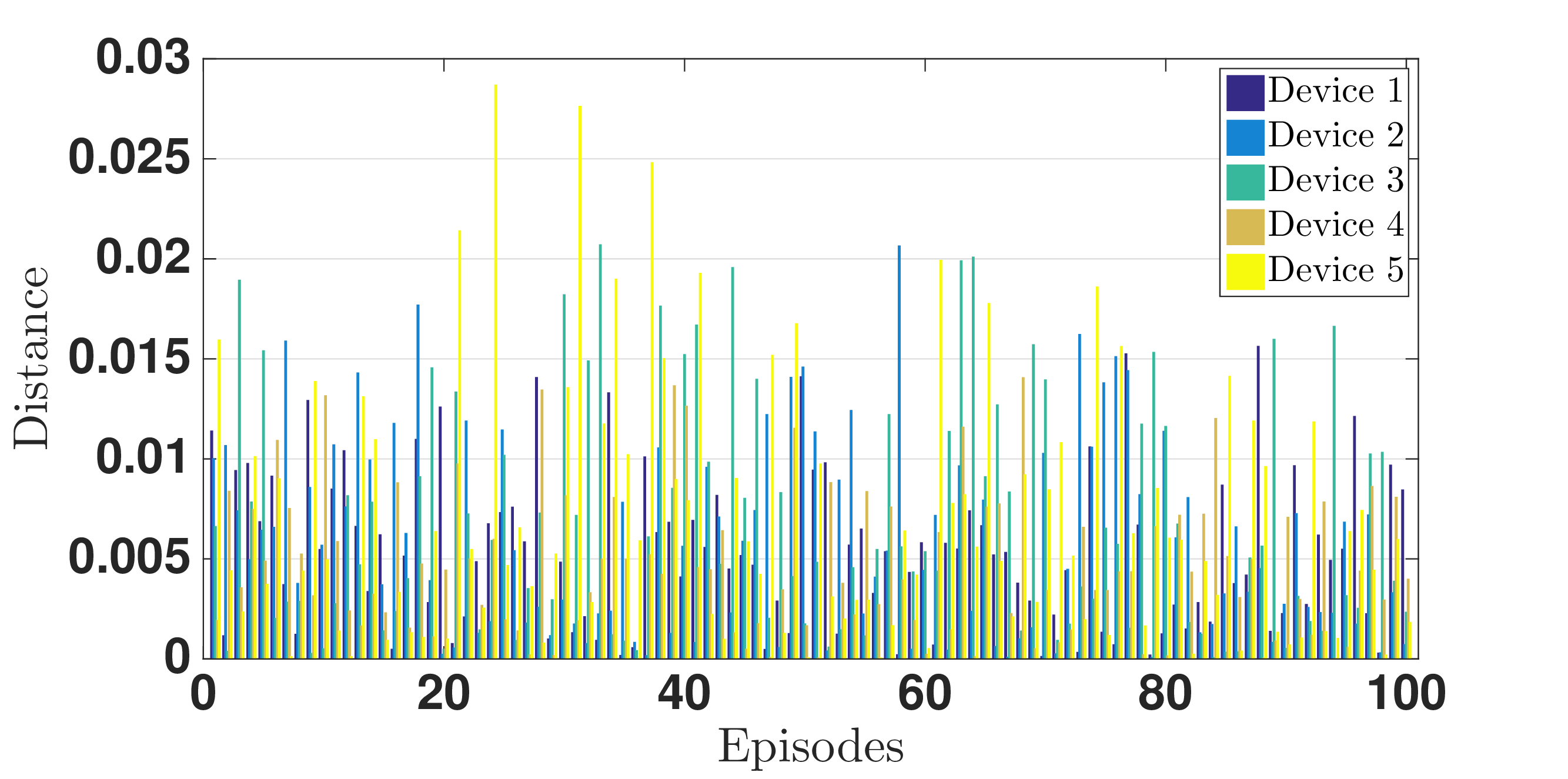} & \includegraphics[width=3in]{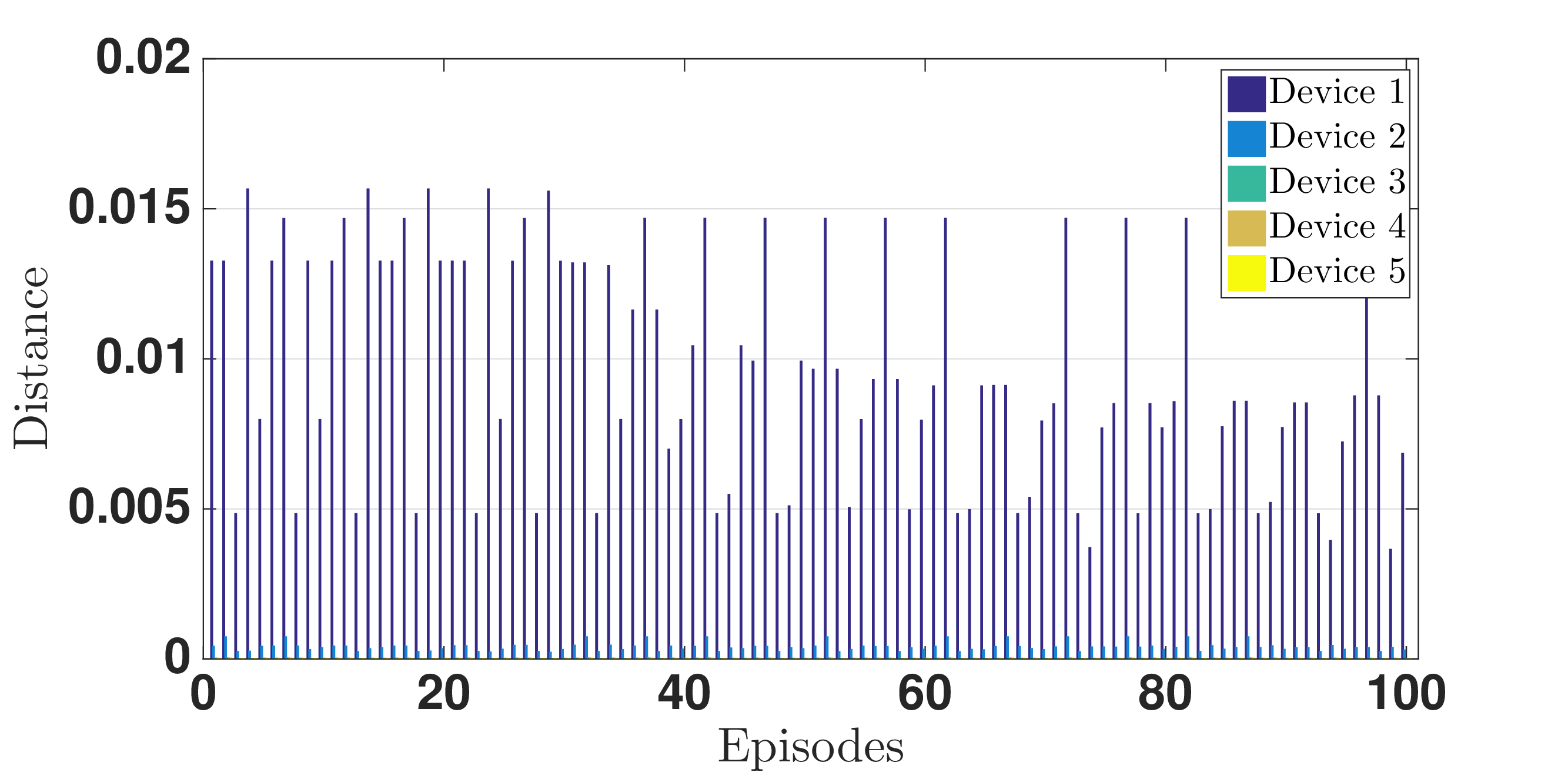} 
\\ (e) The GAE-based attack with CIFAR-10. & (f) The MP attack with CIFAR-10.
\end{tabular}
\end{center}
\caption{Taking FL with five devices as an example, the Euclidean distances between the local models and the global model are presented, where device 1 is the attacker and launches the new GAE-based attack or the MP attack.}
\label{fig_distance_episodes}
\end{figure*}

\subsubsection{Detection of the Attack}
Existing model poisoning attacks on FL aim to maximize the training loss of FL models. One way to detect these malicious attacks is to compare the distances (or differences) between the local models and the global model. A larger distance can be considered as an indication of a malicious local model, and the server can detect it accordingly. Both the Euclidean distance and cosine distance are commonly used metrics to assess the similarities between two vectors. Particularly, the Euclidean distance can assess the conformity between two vectors by capturing both the magnitude and direction of two local models. For this reason, the Euclidean distance is considered in this paper, which is consistent with many recent studies, e.g.,~\cite{zhang2022fldetector} and~\cite{li2021lomar}.

To evaluate the invisibility of the proposed GAE-based, data-agnostic, model poisoning attack Fig.~\ref{fig_distance_episodes} presents the Euclidean distance between the local models and the global model, with device 1 being the attacker. It can be observed that, in general, the local models with the MNIST dataset have the smallest distance compared to the other two datasets. This is expected as the handwritten digits in MNIST are relatively simple to recognize or falsify.

As shown in Figs.~\ref{fig_distance_episodes}(a), \ref{fig_distance_episodes}(c), and \ref{fig_distance_episodes}(e), the Euclidean distances of the malicious local model (i.e., of device 1) generated by the GAE-based attack are below that of the benign local models. This makes it difficult for the aggregator to identify the attacker and defend against the attack.
In contrast, the MP attack results in a significantly larger Euclidean distance between the malicious local model and the global model, making it easier to detect.
This highlights the key advantage of the proposed GAE-based attack, which is designed to generate malicious local models based on the feature correlation between the benign local and global models, making the differences between the malicious local model and the benign local models indistinguishable.

\begin{figure}[htb]
\begin{center}
\includegraphics[width=3in]{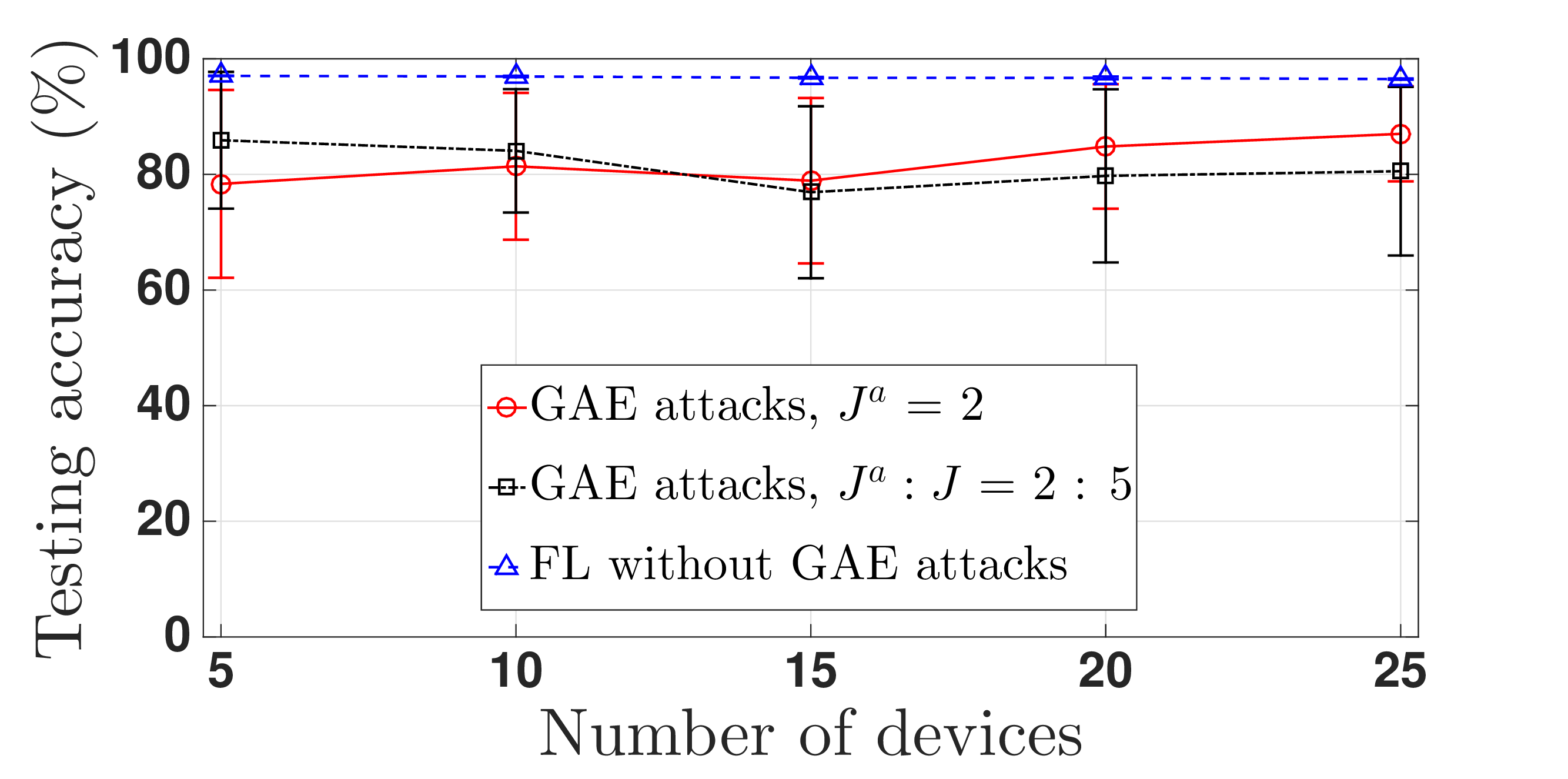} \\
(a) MNIST
\\ \includegraphics[width=3in]{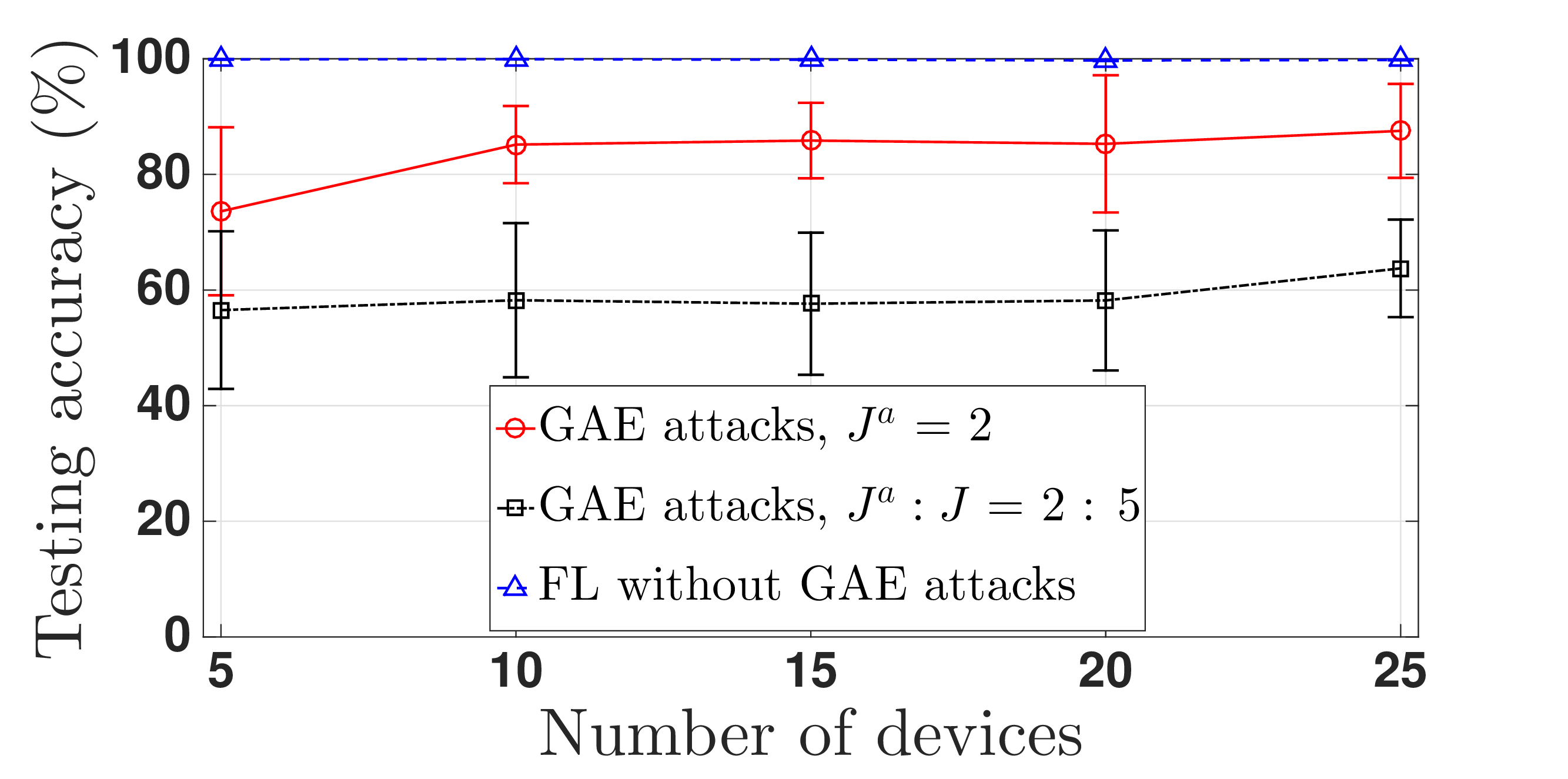} \\
(b) fashionMNIST
\\ \includegraphics[width=3in]{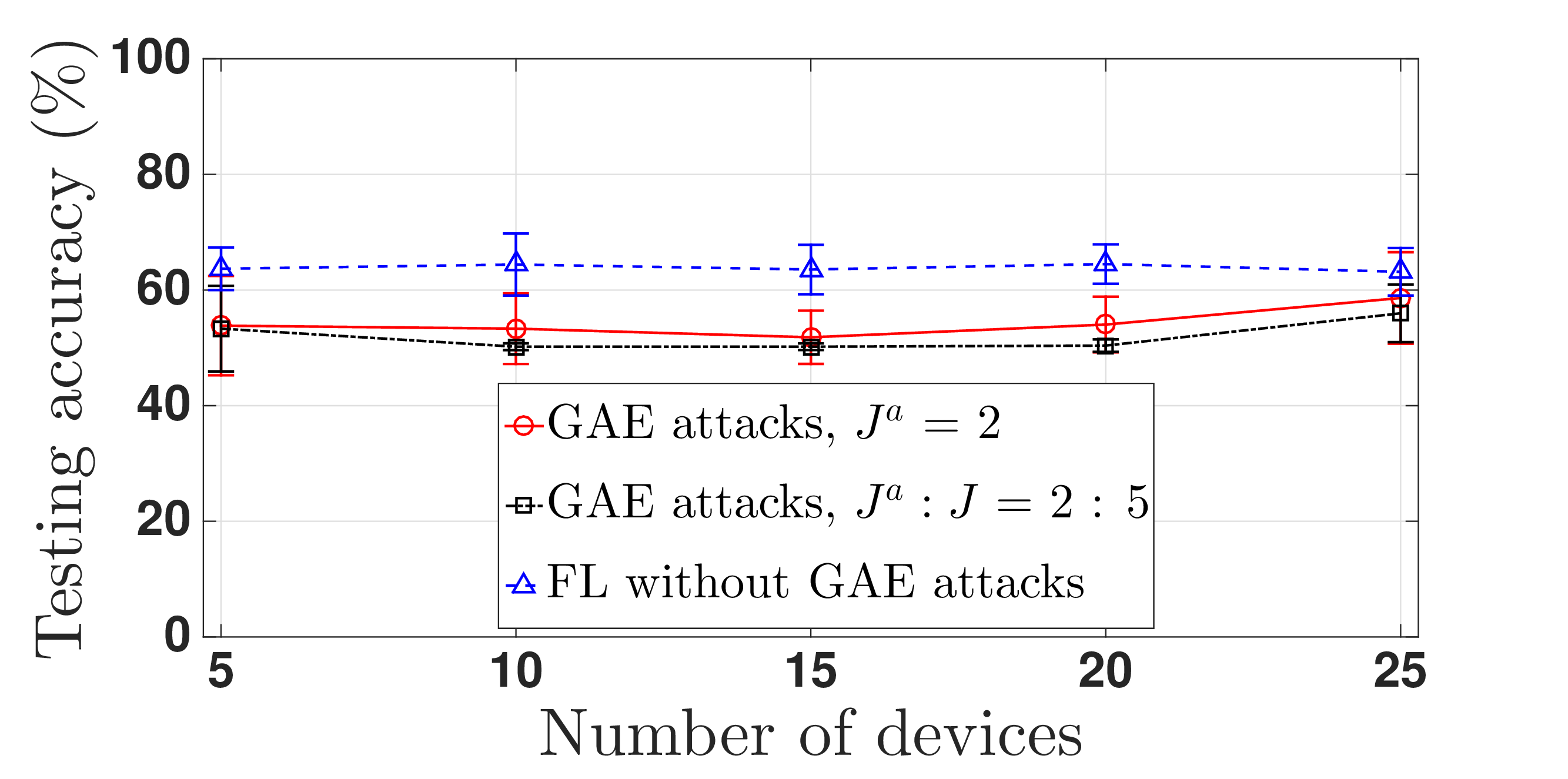} \\
(c) CIFAR-10
\end{center}
\caption{When the number of devices, i.e., $J$, increases from 5 to 25, the $y$-axis shows the average accuracy under the GAE-based attack based on the MNIST, fashionMNIST, or CIFAR-10 datasets. The number of attackers is $J^a=2$ by default. Otherwise, $J^a$ increases with $J$, while keeping the ratio $J^a:J=2: 5$.}
\label{fig_accDevices}
\end{figure}

\subsubsection{Impact of Benign Local Model Number}
Figs.~\ref{fig_accDevices}(a), \ref{fig_accDevices}(b), and \ref{fig_accDevices}(c) show the average accuracy of the local models based on the MNIST, fashionMNIST, and CIFAR-10 datasets, respectively. The total number of benign user devices, i.e., $J$, increases from 5 to 25. The number of attackers is set to $J^a=2$ by default. Otherwise, $J^a$ increases with $J$, while keeping the ratio of $J^a$ to $J$ to be $J^a:J=2: 5$. It is observed that the new GAE-based attack reduces the average accuracy below the performance of FL without the attack. As the number of devices increases, the average accuracy with MNIST, fashionMNIST, and CIFAR-10 drops about 20\%, 37\%, and 12\%, respectively, when $J = 15$. This suggests that the proposed GAE-based attack can effectively infect FL, regardless of the network size.

It is observed in Fig.~\ref{fig_accDevices} that on the three considered datasets, the average accuracy of the FL under attack gradually increases as $J$ grows from 5 to 25, while $J^a$ remains~2. This confirms that increasing the number of benign users improves the resistance of the FL to the attacks. On the other hand, as the ratio of attackers to benign devices, i.e., $J^a:J$, increases, the proposed GAE-based attack can reduce the average accuracy of the attacked model.

\subsubsection{Impact of Eavesdropped Local Models}
\begin{figure}[htb]
\centering
\includegraphics[width=3.5in]{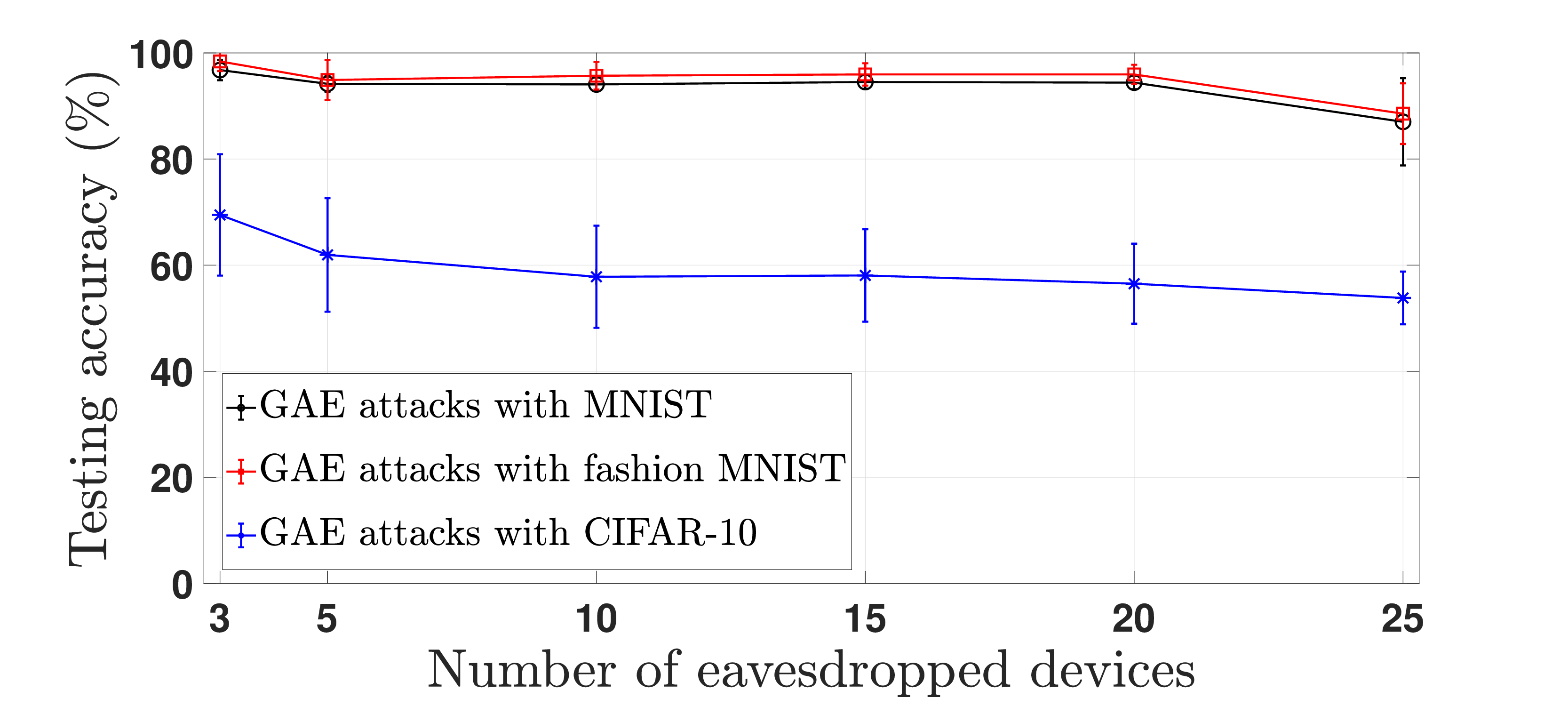}
\caption{The number of eavesdropped benign user devices' $\pmb{\omega}_j(t)$ increases from 3 to 25, based on the MNIST, fashionMNIST, or CIFAR-10 datasets.}
\label{fig_accOverhear}
\end{figure}

Fig.~\ref{fig_accOverhear} plots the average accuracy of the local models under the GAE-based attack based on the MNIST, fashionMNIST, or CIFAR-10 datasets, where the number of benign user devices that the attacker can eavesdrop on increases from 3 to 25. In general, the average accuracy of the local models falls with the growth of the eavesdropped benign user devices. The reason is that overhearing a greater number of benign local models results in capturing more correlation features of the models, leading to the generation of a malicious model for more effective poisoning. The average model accuracy drops substantially by 13.6\%, 11.2\% and 16.4\% on the MNIST, fashionMNIST, and CIFAR-10 datasets, respectively.


\begin{figure}[htb]
\centering
\includegraphics[width=3.5in]{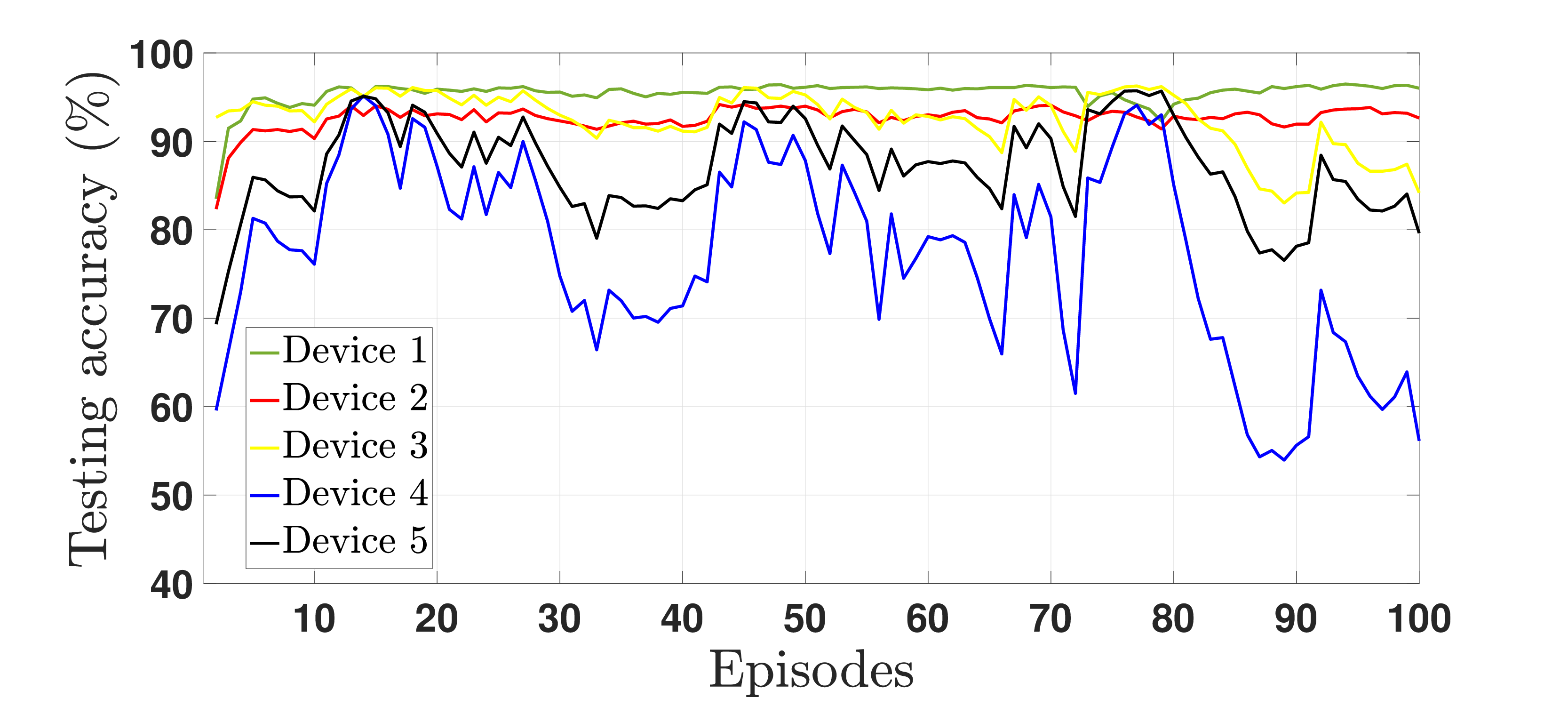} \\
(a) The testing accuracy of local models. \\
\includegraphics[width=3.5in]{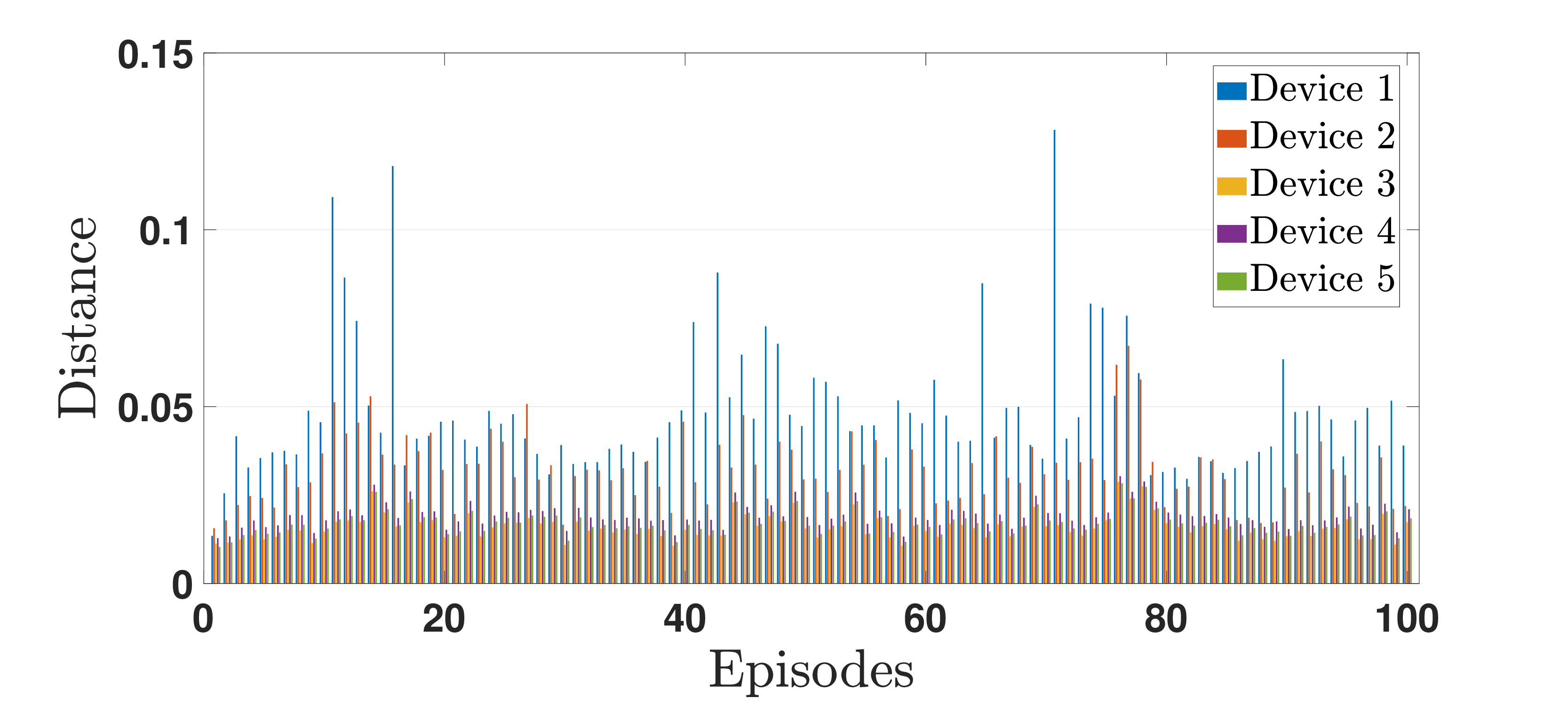} \\
(b) The Euclidean distances between the local models and the global model. 
\caption{The local model testing accuracy and the Euclidean distances under the VAE-based attack.}
\label{fig_accVAE}
\end{figure}

\subsubsection{Compared with Variational Autoencoder (VAE)-based Alternative} 
Fig.~\ref{fig_accVAE} illustrates the effects of the VAE-based attack on FL over 100 communication rounds, where five user devices and the MNIST dataset are considered. Fig.~\ref{fig_accVAE}(a) reveals a consistent variation in local model testing accuracy under the VAE-based attack, with all five devices demonstrating analogous patterns. This is consistent with the observation made on the proposed GAE-based attack in Fig.~\ref{fig_accRounds}. Fig.~\ref{fig_accVAE}(b) shows the Euclidean distances between the local and global models under the VAE-based attack. A striking observation is that the malicious local model (from device 1) constructed via the VAE-based attack possesses a significantly larger Euclidean distance than the benign local models. This suggests that detecting the VAE-based attacks on the server side is feasible by assessing the Euclidean distance. The underlying reason is that the VAEs are general autoencoders and can handle high-dimensional and continuous data, such as images and audio. They do not capture graph structures inside data, as opposed to GAEs.

\section{Conclusion}
\label{sec_cond}
In this paper, we have investigated a new, data-agnostic, model poisoning attack to FL, where the proposed adversarial GAE gives rise to an infection of benign user devices and the FL training accuracy gradually drops. The adversarial GAE allows the attacker to extract the common underlying data features of the benign local models as well as their correlations to generate the malicious model with which the FL training loss is maximized. Since the malicious and benign local models are indistinguishable, it is difficult to identify the GAE-based attack at the server. We implemented the GAE-based attack against the FL on SVM models using PyTorch. Performances were evaluated based on the MNIST, fashionMNIST, and CIFAR-10 datasets. 

\appendices
\section{Proof of Theorem~1}\label{appendix_convergence bound}
Based on the Taylor expansion and the assumption that the gradient of the global loss function is $L$-Lipschitz continuous, it follows that 
\begin{equation}\label{eq_taylor_expansion}
    \begin{aligned}
     F\!({\pmb \omega}^{a}_{g}(t\!+\!1)) \!&-\!F\!({\pmb \omega}^{a}_{g}(t)) \!\leq\!  \nabla F\!\left({\pmb \omega}^{a}_{g}(t) \right) \Big({\pmb \omega}^{a}_{g}(t\!+\!1)
     \\
    &  - {\pmb \omega}^{a}_{g}(t) \Big)\! +\! \frac{L}{2} \!\left\| {\pmb \omega}^{a}_{g}(t\!+\!1) \!- \!{\pmb \omega}^{a}_{g}(t) \right\|^2.
    \end{aligned}
\end{equation} 
By substituting~\eqref{eq_glbAttacks} into~\eqref{eq_taylor_expansion}, we obtain \eqref{eq_F_w_t1} on the top of the next page.
\begin{figure*}
\begin{subequations}\label{eq_F_w_t1}
		\begin{align}
	  F({\pmb \omega}^{a}_{g}(t+1)) -F({\pmb \omega}^{a}_{g}(t)) 
	  &  \leq \nabla F\left({\pmb \omega}^{a}_{g}(t) \right)  \Big[\frac{1}{D} \sum_{j=1}^{J} D_j (\pmb{\omega}_j(t+1) - \pmb{\omega}_j(t)) 
	   + \frac{D_a}{D} (\pmb{\omega}^a(t+1) - \pmb{\omega}^a(t))\Big] \notag \\
	  & \quad + \frac{L}{2}  \Big\|\frac{1}{D} \sum_{j=1}^{J} D_j (\pmb{\omega}_j(t+1) - \pmb{\omega}_j(t)) + \frac{D_a}{D} (\pmb{\omega}^a(t+1) - \pmb{\omega}^a(t))\Big\|^2 \\
	 & =   \frac{\eta^2 L}{2} \bigg\|\frac{1}{D} \sum_{j=1}^{J} D_j(t) \nabla F_j ({\pmb{\omega}}_j(t)) + \frac{D_a}{D} \nabla F_a ({\pmb{\omega}}^{a}(t))\bigg\|^2 \notag \\
	 & \quad -  \eta \nabla F\left({\pmb \omega}^{a}_{g}(t) \right) \left[\frac{1}{D(t)} \sum_{j=1}^{J} D_j(t) \nabla F_j ({\pmb{\omega}}_j(t))  + \frac{D_a}{D} \nabla F_a({\pmb{\omega}}^{a}(t)) \right].
		\end{align}
\end{subequations}
\end{figure*}

By substituting ${\pmb{\omega}}_j(t + 1) = \pmb{\omega}_j(t) - \eta \nabla F_j({\pmb{\omega}}_j(t))$ and $\pmb{\omega}^{a}_{g}(t) =  \sum_{j=1}^{J} \frac{D_j}{D} \pmb{\omega}_j(t) + \frac{D_a}{D} \left(\pmb{\omega}^{a}(t + 1) - \pmb{\omega}^{a}(t)\right) $ into \eqref{eq_F_w_t1}, the expectation of $F(\pmb{\omega}^{a}_{g}(t+1))$ can be given by
\begin{subequations}\label{eq-expected-loss}
\begin{align}
     \mathbb{E}\!&\left[F(\pmb{\omega}^{a}_{g}(t\!+\!1)) \right] \!\leq\! F\!(\pmb{\omega}^{a}_{g}(t)) \!- \!\eta \!\left\|\nabla \! F\!\left({\pmb \omega}^{a}_{g}(t) \right) \right\|^2\!+\nonumber\\
     &~~ \frac{\eta^2 L}{2}\! \mathbb{E} \bigg[\bigg\|\! \sum_{j=1}^{J} \!\frac{D_j}{D}\!\nabla \!F_j ({\pmb{\omega}}_j(t)) \!+\! \frac{D_a}{D} \!\nabla \!F_J ({\pmb{\omega}}^{a}(t))\bigg\|^2\bigg].\nonumber
\end{align}
\end{subequations}

It is generally assumed in FL that $$\mathbb{E}\left[\sum_{j=1}^{J} \frac{D_j}{D} \! \nabla \! F_j ({\pmb{\omega}}_j(t)) \!+ \!\frac{D_a}{D} \! \nabla \! F_j ({\pmb{\omega}}^{a}(t))\right] \!= \! \nabla \! F({\pmb \omega}^{a}_{g}(t)).$$

According to Jensen's inequality, $\mathbb{E}(x^2) \leq \mathbb{E}^2(x)$. Then,
\begin{equation}
\begin{aligned}
    &\mathbb{E}\left[ \left\|\! \sum_{j=1}^{J} \frac{D_j}{D}\! \nabla \!F_j ({\pmb{\omega}}_j(t)) \!+ \!\!\frac{D_a}{D}\! \nabla \! F_j ({\pmb{\omega}}^{a}(t))\right\|^2\right] \!\!\!\leq \!\!\left\|\nabla\! F\!\!\left({\pmb \omega}^{a}_{g}(t)\! \right)\! \right\|^2.
\end{aligned}    
\end{equation}
As a result, $\mathbb{E}\left[F(\pmb{\omega}^{a}_{g}(t+1)) \right]$ is bounded by
\begin{subequations}\label{eq_Fw1_Fw}
\begin{align}
   & \mathbb{E}\left[F(\pmb{\omega}^{a}_{g}(t+1)) \right] \notag \\
     & \leq \mathbb{E}\left[ F(\pmb{\omega}^{a}_{g}(t))\right] 
    + (\frac{\eta^2 L}{2} - \eta)\left\|\nabla F\left({\pmb \omega}^{a}_{g}(t) \right) \right\|^2\\
    & \overset{\eta = \frac{1}{L}}{=} \mathbb{E}\left[ F(\pmb{\omega}^{a}_{g}(t)) \right] - \frac{\eta}{2}\left\|\nabla F\left({\pmb \omega}^{a}_{g}(t) \right) \right\|^2.
\end{align}
\end{subequations}

Next, we derive the relationship between $\nabla F({\pmb \omega}^{a}_{g}(t))$ and $\nabla F({\pmb \omega}_{g}(t))$. Since $d(\pmb{\omega}_j^a, \pmb{\omega}_{g}^a) = \|\pmb{\omega}_j^a - \pmb{\omega}_{g}^a\| \leq d_{T}$, we have
\begin{subequations}\label{eq-dis}
    \begin{align}
    &\!\!\! \|\pmb{\omega}^a (t\!+\!1) \!- \! \pmb{\omega}^a (t)\| \notag\\
    & \leq \!  \|\pmb{\omega}^a (t\!+\!1)\! -\! \pmb{\omega}_{g}^a (t\! + \!1)\| \! + \! \|\pmb{\omega}_{g}^a (t\!+\!1) \!-\! \pmb{\omega}^a (t)\| \notag\\
    & = \!d_{T} \!+\! \|\pmb{\omega}_{g}^a (t\!+\!1)\! - \!\pmb{\omega}^a (t)\| \notag\\
    & \leq \!d_{T} \!\!+\!\! \|\pmb{\omega}_{g}^a (t\!+\!1) \!-\! \pmb{\omega}_{g}^a (t)\| \!\!+\!\! \|\pmb{\omega}_{g}^a (t) \!-\! \pmb{\omega}^a (t)\| \label{eq: 25}\\
    & = 2 d_{T} + \|\pmb{\omega}_{g}^a (t+1) - \pmb{\omega}_{g}^a (t)\|,
    \end{align}
\end{subequations}
where \eqref{eq: 25} is based on the triangle inequality.

Likewise, we also have
\begin{subequations}\label{eq-delta-blb 1}
    \begin{align}
        & \|\pmb{\omega}_{g}^a (t+1) - \pmb{\omega}_{g}^a (t)\| \notag \\
        & =\! \Big\|\!\big(\pmb{\omega}_{g}(t\!+\!1) \!+ \!\frac{D_a}{D}\pmb{\omega}^a(t\!\!+\!\!1) \! \big) \!\!-\!\! \big(\pmb{\omega}_{g}(t) \!+\! \frac{D_a}{D}\pmb{\omega}^a(t)\! \big)\Big\| \\
        & \leq \! \left\|\pmb{\omega}_{g}(t\!+\!1)\! -\!\pmb{\omega}_{g}(t) \right\| \!+\! \frac{D_a}{D}\left\| \pmb{\omega}^a(t\!+\!1)\! -\! \pmb{\omega}^a(t)\right\| \\
       & \leq \left\|\pmb{\omega}_{g}(t+1) -\pmb{\omega}_{g}(t) \right\| \notag \\
       & \;\;\;+ \frac{D_a}{D} \left( 2 d_{T} \!+\! \|\pmb{\omega}_{g}^a (t+1) - \pmb{\omega}_{g}^a (t)\| \right). \label{eq: 26c}
    \end{align}
\end{subequations}
By reorganizing \eqref{eq: 26c}, it follows that
\begin{subequations}\label{eq-delta-blb 2}
\begin{align}
     \|\pmb{\omega}_{g}^a (t+1)  &- \pmb{\omega}_{g}^a (t)\| \leq\frac{D}{D-D_a} \times  \\
    &  \Big\|\pmb{\omega}_{g}(t\!+\!1)\!-\!\pmb{\omega}_{g}(t) \Big\| \!+ \!\frac{2D_a d_{T}}{D-D_a}.
\end{align}
\end{subequations}
By substituting \eqref{eq-delta-blb 2} into \eqref{eq-dis}, it follows that
\begin{subequations}\label{eq-deltaWJ}
    \begin{align}
     \|\pmb{\omega}^a (t+1)& -  \pmb{\omega}^a (t)\| \leq \frac{D}{D-D_a}\times \\
    & \left\|\pmb{\omega}_{g}(t+1) -\pmb{\omega}_{g}(t) \right\| + \frac{2D d_{T}}{D-D_a}.
    \end{align}
\end{subequations}
By taking expectation on both sides of \eqref{eq-delta-blb 2}, we have
\begin{equation}\label{eq-delta-blb 3}
\begin{aligned}
     \mathbb{E} [ \|\pmb{\omega}_{g}^a &(t+1) - \pmb{\omega}_{g}^a (t)\| ] \leq \frac{D}{D-D_a}\times \\
    &  \mathbb{E} [\left\|\pmb{\omega}_{g}(t+1) -\pmb{\omega}_{g}(t) \right\|] + \frac{2D_a d_{T}}{D-D_a}.
\end{aligned}
\end{equation}
Note that $\mathbb{E} [\|\pmb{\omega}_{g}^a (t+1) - \pmb{\omega}_{g}^a (t)\|] = \eta \|\nabla F(\pmb{\omega}_{g}^a (t)) \|$ and $\mathbb{E} [\|\pmb{\omega}_{g} (t+1) - \pmb{\omega}_{g} (t)\|] = \eta \|\nabla F(\pmb{\omega}_{g} (t)) \|$. By substituting them into both sides of \eqref{eq-delta-blb 3} and then reorganizing \eqref{eq-delta-blb 3}, we obtain
\begin{equation}\label{eq-delta_glb_final}
\begin{aligned}
    \|\nabla F(\pmb{\omega}_{g}^a (t)) \|\!
    & \leq \!\!\frac{D}{D\!\!-\!\!D_a} \|\nabla F(\pmb{\omega}_{g} (t)) \|\! + \!\frac{2D_a d_{T}}{(D-D_a)\eta}.
\end{aligned}
\end{equation}

By substituting \eqref{eq-delta_glb_final} into \eqref{eq_Fw1_Fw}, it follows 
\begin{subequations}\label{eq_Fw1_Fw 2}
\begin{align}
   & \mathbb{E}\left[F(\pmb{\omega}^{a}_{g}(t+1)) \right]-\mathbb{E}\left[ F(\pmb{\omega}^{a}_{g}(t))\right] \notag \\
   & \leq - \frac{\eta}{2}\left\|\nabla \! F\left({\pmb \omega}^{a}_{g}(t) \right) \right\|^2\\
    & \leq - \frac{\eta}{2} \left(\frac{D}{D\!-\!D_a} \|\nabla F(\pmb{\omega}_{g} (t)) \| \!+\! \frac{2D_a d_{T}}{(D-D_a)\eta} \right)^2\\
    & \leq - \frac{\eta D^2}{2(D-D_a)^2} \|\nabla F(\pmb{\omega}_{g} (t)) \|^2.
\end{align}
\end{subequations}

Considering the Polyak-Lojasiewicz condition,
we have
\begin{equation}\label{eq_delta_Fw}
\begin{aligned}
    \|\nabla F(\pmb{\omega}_{g}(t) \|^2 \geq 2 \rho \left( F(\pmb{\omega}_{g}(t) - F(\pmb{\omega}_g^{\ast})\right).
\end{aligned}     
\end{equation}
Given the convex loss function of SVM models, it follows:  
\begin{equation}\label{eq_convex}
    \begin{aligned}
    F({\pmb \omega}^{a}_{g}(t)) &\leq  F({\pmb \omega}_{g}\left(t)\right) + F \left(\frac{D_a}{D}\pmb{\omega}^a(t)\right)\\
    &\leq  F({\pmb \omega}_{g}\left(t)\right) + \frac{D_a}{D} F \left(\pmb{\omega}^a(t)\right).
    \end{aligned}
\end{equation}

By substituting \eqref{eq_delta_Fw} and \eqref{eq_convex} into \eqref{eq_Fw1_Fw 2}, we have
\begin{subequations}\label{eq_Fw1_Fw 3}
\begin{align}
    & \mathbb{E}\left[F(\pmb{\omega}^{a}_{g}(t+1)) \right]-\mathbb{E}\left[ F(\pmb{\omega}^{a}_{g}(t))\right] \notag \\
    & \leq - \frac{\eta \rho D^2}{(D-D_a)^2} \left( F(\pmb{\omega}_{g}(t) - F(\pmb{\omega}_g^{\ast})\right) \\
    & \leq  \! \! - \frac{\eta \rho D^2}{(D \! \!- \! \!D_a)^2} \left(  \!\! F(\pmb{\omega}^{a}_{g}(t) \!- \! \frac{D_a}{D} F \! \left(\pmb{\omega}^a(t)\right) \! -  \!F(\pmb{\omega}_g^{\ast})   \!\!\right).
\end{align}
\end{subequations}
By restructuring~\eqref{eq_Fw1_Fw 3}, we have
\begin{equation}\label{eq-recurrence}
\begin{aligned}
& \mathbb{E}\left[  F({\pmb \omega}^a_{g}\left(t + 1)\right)\right] - F({\pmb \omega}_g^*)\\
& \leq  \left(1 - \frac{\rho \eta D^2}{(D-D_a)^2}\right)  \left(\left[F({\pmb \omega}^a_{g}(t) )\right] - F({\pmb \omega}_g^*)\right) \\
& \quad + \frac{\rho \eta D D_a}{(D-D_a)^2}  F \left(\pmb{\omega}^a(t)\right)\\
& \leq  \left(1 - \zeta\right)  \left(\left[F({\pmb \omega}^a_{g}(t) )\right] - F({\pmb \omega}_g^*)\right)  + \frac{\rho \eta D D_a}{(D-D_a)^2} F^{\max},
\end{aligned}
\end{equation}
where $\zeta = 1 - \frac{\rho \eta D^2}{(D-D_a)^2}$, and the second inequality is due to the constrained problem in~\eqref{eq_opt_obj} and~\eqref{eq_const_dist} has a maximum value, i.e., $ F (\pmb{\omega}^a(t)) \leq F^{\max}$.

Finally, applying mathematical induction upon~\eqref{eq-recurrence} gives
\begin{equation}
\begin{aligned}
& \mathbb{E}\left[ F({\pmb \omega}^{a}_{g}(t) )\right] - F({\pmb \omega}_g^*) \\
& \leq \left[ \mathbb{E}\left[F({\pmb \omega}_g(0))\right] - F({\pmb \omega}_g^*)\right] \zeta^t + \frac{1-\zeta^t}{1-\zeta} \cdot \frac{\rho \eta D D_a}{(D-D_a)^2} F^{\max}\\
& = \Theta  \zeta^t + \frac{1-\zeta^t}{1-\zeta} \cdot \frac{\rho \eta D D_a}{(D-D_a)^2} F^{\max},
\end{aligned}
\end{equation}
which concludes this proof.


\ifCLASSOPTIONcaptionsoff
  \newpage
\fi

\bibliographystyle{IEEEtran}
\bibliography{bibFLGAE}

\end{document}